\documentclass[11pt]{article}

\usepackage[preprint]{acl}
\usepackage{times}
\usepackage{latexsym}
\usepackage{amssymb}
\usepackage[T1]{fontenc}
\usepackage[utf8]{inputenc}
\usepackage{microtype}
\usepackage{inconsolata}

\usepackage{graphicx}
\usepackage{hyperref}
\usepackage{url}
\usepackage{graphicx}
\usepackage[table,xcdraw]{xcolor}
\usepackage{colortbl}
\usepackage{tabularx}
\usepackage{booktabs}
\usepackage{multirow}
\usepackage{wrapfig}
\usepackage{caption} 
\usepackage{enumitem}
\usepackage{amsmath}
\usepackage{array}
\usepackage{tikz}
\usepackage{listings}
\usepackage{xcolor}
\usepackage{ragged2e}
\usepackage[most]{tcolorbox}
\usepackage{needspace}
\usepackage{enumitem}
\usepackage{hyperref}
\newcolumntype{L}[1]{>{\RaggedRight\arraybackslash}p{#1}} 
\usepackage{graphicx}
\usepackage{subcaption}
\usepackage{caption}
\usepackage{float}

\usepackage[utf8]{inputenc}
\usepackage[T1]{fontenc}   
\usepackage{hyperref}       
\usepackage{url}         
\usepackage{booktabs}    
\usepackage{amsfonts}   
\usepackage{nicefrac}     
\usepackage{microtype}      
\usepackage{xcolor}        
\usepackage[utf8]{inputenc}
\usepackage[T1]{fontenc}
\usepackage{ifthen}

\usepackage{hyperref}
\usepackage{url}
\usepackage{booktabs}
\usepackage{amsfonts}
\usepackage{nicefrac}
\usepackage{microtype}
\usepackage{graphicx}
\usepackage{array}
\usepackage{ifthen}     
\usepackage{tocloft}

\renewcommand{\addcontentsline}[3]{}

\usepackage{makecell}
\usepackage{listings} 
\usepackage{xcolor}   

\tcbset{
  failurebox/.style={
    enhanced,
    breakable,
    colback=white,
    colframe=black,
    fonttitle=\bfseries,
    boxrule=0.5pt,
    arc=1.5pt,
    left=4pt,
    right=4pt,
    top=4pt,
    bottom=4pt,
    before skip=4pt,
    after skip=6pt,
  }
}

\lstdefinestyle{leanstyle}{
    basicstyle=\ttfamily\footnotesize,
    breaklines=true,
    columns=fullflexible,
    keepspaces=true,
    showspaces=false,
    showstringspaces=false,
    showtabs=false,
    numbers=none,
    frame=none,
    backgroundcolor=\color{white},
    aboveskip=0pt,
    belowskip=0pt
}

\newcounter{example}
\newcommand{\exampletitle}[1]{%
  \refstepcounter{example}%
  Example~\theexample: #1%
}

\definecolor{codegreen}{rgb}{0,0.6,0}
\definecolor{codegray}{rgb}{0.5,0.5,0.5}
\definecolor{codepurple}{rgb}{0.58,0,0.82}
\definecolor{backcolour}{rgb}{0.95,0.95,0.92}

\lstdefinestyle{mystyle}{
    backgroundcolor=\color{backcolour},
    commentstyle=\color{codegreen},
    keywordstyle=\color{magenta},
    numberstyle=\tiny\color{codegray},
    stringstyle=\color{codepurple},
    basicstyle=\ttfamily\footnotesize,
    breakatwhitespace=false,
    breaklines=true,
    captionpos=b,
    keepspaces=true,
    numbers=left,
    numbersep=5pt,
    showspaces=false,
    showstringspaces=false,
    showtabs=false,
    tabsize=2,
    language=Python
}
\usepackage[T1]{fontenc}
\usepackage{microtype}
\usepackage{xurl}
\usepackage{listings}
\usepackage{tabularx}
\usepackage{array}
\usepackage{booktabs}
\usepackage{seqsplit}
\usepackage{upquote}

\newcommand{\icode}[1]{\texttt{\seqsplit{#1}}}

\newcolumntype{Y}{>{\raggedright\arraybackslash}X}

\lstdefinestyle{reprolisting}{
  basicstyle=\ttfamily\scriptsize,
  breaklines=true,
  breakatwhitespace=false,
  columns=fullflexible,
  keepspaces=true,
  showstringspaces=false,
  upquote=true,
  frame=single,
  framerule=0.2pt,
  xleftmargin=0.6em,
  xrightmargin=0.6em,
  aboveskip=0.6em,
  belowskip=0.6em
}

\lstdefinestyle{promptstyle}{
  style=reprolisting
}

\lstdefinestyle{leanstyle}{
  style=reprolisting
}

\lstdefinestyle{jsonstyle}{
  style=reprolisting
}

\title{RePro: Proof-Verified Benchmark Rewriting for Reliable Evaluation of LLM Mathematical Problem Solving}

\author{
\textbf{Xiyuan Zhou\textsuperscript{1}\thanks{Equal contribution.}},
\textbf{Zhuoqi Li\textsuperscript{2}\footnotemark[1]},
\textbf{Xinlei Wang\textsuperscript{3}},
\textbf{Yirui He\textsuperscript{2,4}},
\textbf{Yuhao Wu\textsuperscript{2}},
\textbf{Yuheng Cheng\textsuperscript{2}},  \\
\textbf{Yan Xu\textsuperscript{1}\thanks{Corresponding authors.}},
\textbf{Junhua Zhao\textsuperscript{2,5}\footnotemark[2]},
\textbf{Jinjin Gu\textsuperscript{3}\footnotemark[2]}\vspace{2mm}
\\
\textsuperscript{1}Nanyang Technological University,
\textsuperscript{2}The Chinese University of Hong Kong, Shenzhen,\\
\textsuperscript{3}INSAIT, Sofia University ``St. Kliment Ohridski'',
\textsuperscript{4}Shenzhen Loop Area Institute,
\textsuperscript{5}AIRS
\\
\texttt{xiyuan002@e.ntu.edu.sg, zhuoqili1@link.cuhk.edu.cn} \\
\texttt{xinlei.wang@insait.ai, yiruihe@link.cuhk.edu.cn} \\
\texttt{yuhaowu@link.cuhk.edu.cn, yuhengcheng@link.cuhk.edu.cn} \\
\texttt{xuyan@ntu.edu.sg, zhaojunhua@cuhk.edu.cn, jinjin.gu@insait.ai}
}

\begin{document}
\maketitle

\begin{abstract}

Data contamination undermines the reliable evaluation of large language models (LLMs) on mathematical problem solving. While rewriting-based evaluation mitigates memorization, existing methods lack guarantees of problem validity and answer correctness. We propose Proof-Verified Benchmark Rewriting (RePro), the first framework to integrate Lean-oriented neural automated theorem provers (ATPs) into benchmark rewriting, which rewrites problems and regenerates answers with correctness ensured by Lean-verified proofs. Experiments on GSM8K and MATH show that RePro's retained rewritten instances achieve 100\% well-definedness, feasibility, and answer correctness, while existing methods still produce invalid or incorrect instances. Moreover, several models exhibit accuracy drops on proof-verified rewritten benchmarks, suggesting that their performance is sensitive to surface-level and structural variations and may partly reflect memorization effects. Our source code and data are available at \url{https://github.com/AI4Engi/RePro}.

\end{abstract}

\section{Introduction}

Evaluating mathematical capability is essential for understanding the reasoning abilities of large language models (LLMs) \cite{shao2024deepseekmath,ahn-etal-2024-large}. However, benchmark reliability is challenged by data contamination, as training corpora and evaluation benchmarks often share public sources \cite{chen-etal-2025-benchmarking-large,cheng2025survey}. Such overlap may allow models to achieve high scores through memorization rather than genuine reasoning \cite{li-etal-2024-open-source, zhou2026engibench, zhao-etal-2025-multimodal}. Recent dynamic evaluation methods, including benchmark rewriting, interactive evaluation, and multi-agent evaluation, aim to reduce contamination \cite{chen-etal-2025-benchmarking-large}. However, their reliance on heuristic rewriting or model-generated processes makes it difficult to guarantee problem validity and answer correctness.

\begin{figure}[]
\centering
\includegraphics[width=0.5\textwidth]{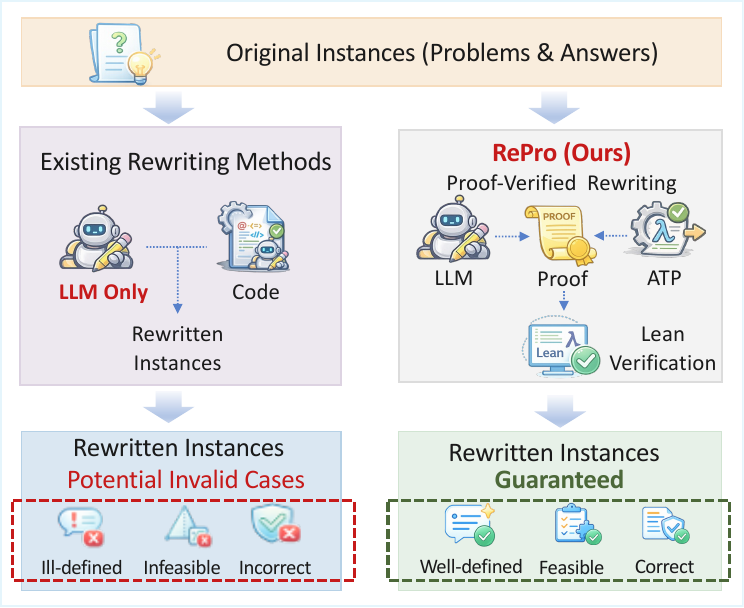}  %
\caption{Overview of RePro. 
Existing rewriting methods may produce invalid problems or incorrect answers. RePro integrates formal verification to ensure that rewritten instances are valid questions and paired with verified answers, enabling reliable LLM evaluation.
}
\label{fig: intro figure}
\end{figure}

Benchmark reliability remains a concern in existing evaluations, even for influential expert benchmarks such as GPQA \cite{rein2024gpqa} and HLE \cite{phan2025lastexam}, which have advanced frontier LLM evaluation. HLE-Verified further highlights the importance of answer reliability, reporting that within HLE's mathematical category, problem validity exceeds 92\% while answer validity is 59.6\% \cite{zhai2026hle}.
This suggests that benchmark reliability depends on both problem validity and answer correctness, reflecting a broader emphasis on verifier-guided reliability in LLM systems \cite{wang2026memguardpersistingverifiersignals}. Accordingly, RePro focuses on mathematical and formally verifiable problems, and evaluates rewritten instances by whether they are well-defined, feasible, and paired with a correct reference answer (see Sec.~\ref{sec: Evaluation Metrics}).

To improve benchmark rewriting reliability, we introduce deterministic proof verification by incorporating Lean-oriented neural automated theorem provers (ATPs) and proof-assistant checking into the rewriting pipeline, replacing heuristic LLM-based evaluation with machine-verifiable reasoning. In RePro, proof search relies on Lean-oriented neural ATPs such as Goedel-Prover \cite{lin2025goedel} and DeepSeek-Prover \cite{ren2025deepseek}. Given a formalized statement, these models generate Lean proof scripts, which are treated as candidate proofs and accepted only after Lean kernel-level verification. Unlike classical ATPs and SMT solvers such as Vampire \cite{kovacs2013first} and Z3 \cite{de2008z3}, which return sound results within supported logical fragments, neural ATPs may generate scripts with compilation failures, target mismatches, or tactic-level errors. RePro therefore retains only proofs that pass Lean kernel-level checking \cite{de2015lean}.

Building on the guarantees provided by formally verified proofs, we propose RePro (Proof-Verified Benchmark Rewriting), a framework for constructing mathematically rigorous rewritten benchmarks. As illustrated in Fig. \ref{fig: intro figure}, in RePro, LLMs generate diverse rewritten problems and perform conservative semantic screening, while ATPs search for candidate proofs and proof assistants verify them. In this way, the rewritten benchmark maintains high diversity while providing verifiable correctness guarantees. Only instances whose reference answers have a formally verified proof are retained, ensuring that the released benchmarks contain only problems with formally verified answers. Detailed methodology is presented in Sec.~\ref{sec: Verifier-in-the-Loop Perturbation}.

Empirical results show that RePro significantly improves the reliability of rewriting-based evaluation. Compared with existing methods, RePro achieves 100\% well-definedness, feasibility, and answer correctness among retained rewritten instances on both GSM8K \cite{cobbe2021gsm8k} and MATH \cite{hendrycks2021measuring}, while prior methods still produce invalid problems or incorrect reference answers. 

Our contributions can be summarized as follows:
(1) We propose RePro, the first benchmark rewriting framework that integrates ATPs and Lean into a unified verification pipeline, retaining only instances with verified reference answers while enforcing problem validity through a three-stage verification process.
(2) We introduce reliability-oriented evaluation criteria for rewritten benchmarks, covering well-definedness, feasibility, and answer correctness. 
(3) We use proof-verified rewriting to analyze reformulation sensitivity and identify potential memorization-related signals.

\begin{figure*}[]
\centering
\includegraphics[width=1\textwidth]{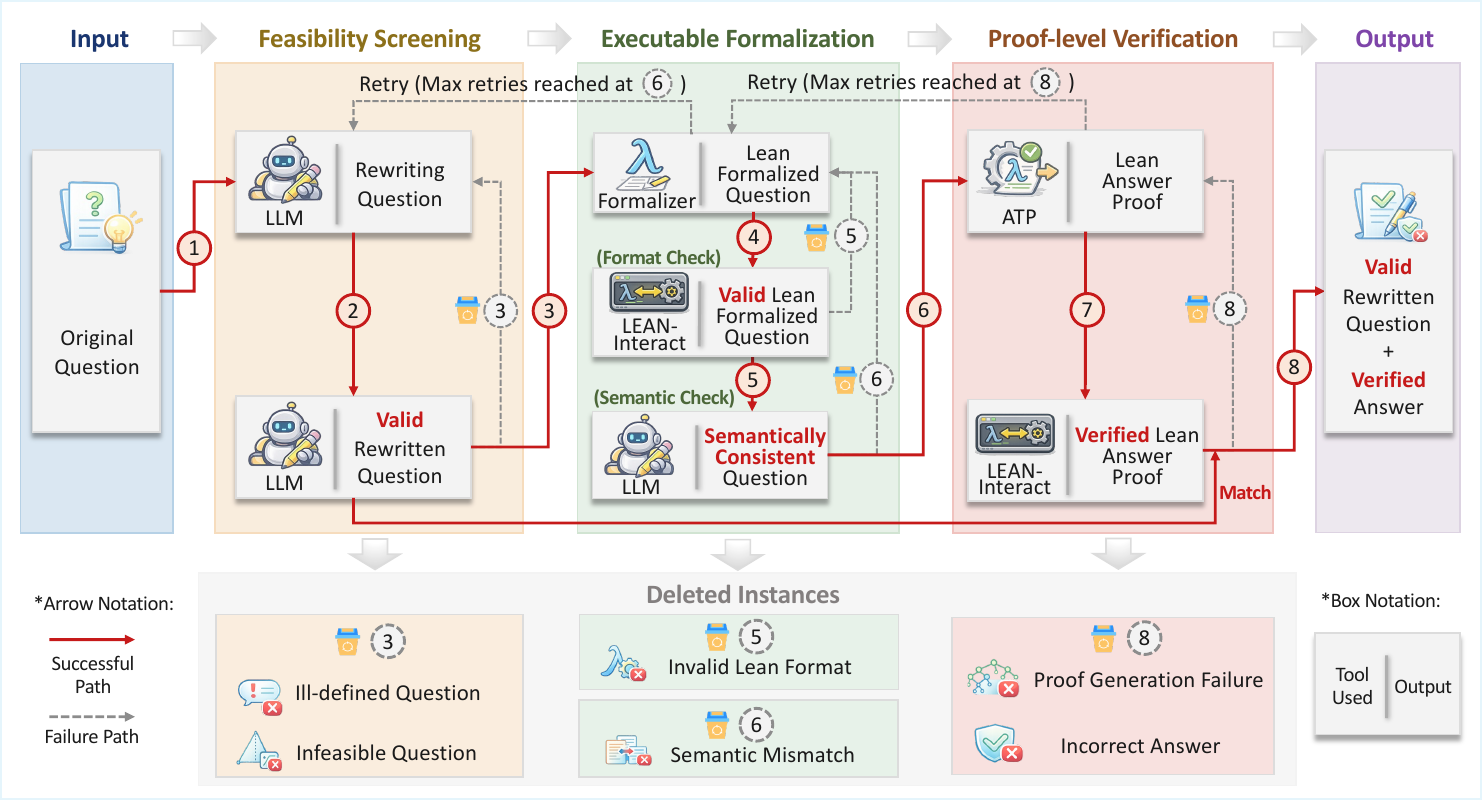}  %
\caption{Framework of the proposed verification pipeline for rewriting-based evaluation.  The pipeline progressively filters rewritten instances to obtain valid questions with verified answers.
}
\label{fig: Framework}
\end{figure*}

\section{Related Work}

\noindent \textbf{Dynamic Benchmark Generation.}\quad
Dynamic benchmark methods mitigate data contamination and expand evaluation coverage by automatically generating new test instances from existing benchmarks. A common approach is benchmark rewriting, which applies semantic or structural transformations, such as synonym paraphrasing \cite{ying2024automating,zhu-etal-2024-inference}, numerical substitution \cite{qian-etal-2024-varbench}, and structural perturbation \cite{cao-etal-2024-structeval}, to weaken memorization cues while reusing existing evaluation resources. Another line of work adopts multi-agent or solver-based frameworks for benchmark construction, such as Benchmark Self-Evolving \cite{wang-etal-2025-benchmark}, BenchAgents \cite{butt2024benchagents}, and the CSP-based logic puzzle benchmark ZebraLogic \cite{lin2025zebralogic}. Despite improving diversity and coverage, these methods still largely rely on heuristic validation, including human inspection, LLM-as-a-Judge, or agent-based checking. Such validation may still leave semantic drift, incorrect labels, or implicit-assumption violations, limiting deterministic reliability guarantees, see Sec.~\ref{sec: Evaluation Metrics} and Sec.~\ref{sec: Rewriting Quality Comparison}.

\noindent \textbf{Formal Verification and Automated Theorem Proving.}\quad
Formal reasoning represents mathematical statements and proofs in a machine-verifiable format, enabling rigorous verification of logical correctness. Proof assistants such as Lean \cite{de2015lean} and Coq \cite{bertot2013interactive} provide formal languages for mathematical reasoning and verify proofs through kernel-level checking. Large formal mathematical libraries such as mathlib support large-scale formalization and automated reasoning \cite{yang2023leandojo}. Classical ATPs and SMT solvers, such as Vampire \cite{kovacs2013first} and Z3 \cite{de2008z3}, solve formal logical problems within supported logics, while hammer systems such as LeanHammer bridge Lean with external provers \cite{zhu2025premise}. In contrast, recent neural Lean provers, including DeepSeek-Prover \cite{ren2025deepseek} and Goedel-Prover \cite{lin2025goedel}, generate candidate Lean proof scripts that must be checked by Lean before acceptance. RePro operates in this Lean/mathlib setting and uses neural Lean provers with kernel-level verification to ensure answer correctness. While prior work mainly studies theorem proving itself, benchmark verification remains underexplored.

\section{RePro} \label{sec: Verifier-in-the-Loop Perturbation}

\subsection{Overview}

To construct rewritten benchmarks with formally verified reference answers, we propose RePro.
Existing rewriting-based approaches typically rely on heuristic validation mechanisms, such as LLM-as-a-Judge or rule-based checking, which cannot provide deterministic guarantees on problem validity or answer correctness. 
To address this limitation, RePro integrates ATPs into the benchmark rewriting process, enabling the verification of reference answers through formal proofs while preserving the diversity of rewritten instances.

RePro follows a progressive verification paradigm. It first prompts an LLM to generate candidate rewrites through numerical reparameterization, logical restructuring, constraint modification, and contextual reconstruction, and filters out instances that fail basic problem-validity checks, such as those with ambiguous statements, missing constraints, or infeasible solutions. The remaining candidates are then translated into executable formal specifications in Lean, providing precise and machine-checkable representations of the rewritten problems. Finally, an automated theorem prover (ATP) performs proof search to generate candidate proofs, whose correctness is checked by the Lean proof assistant. Through this staged process, RePro retains only rewritten instances that are well-defined and feasible and whose answers are supported by formally verified proofs. The prompt templates and implementation details are provided in Appendix~\ref{app:prompt_templates}, and examples of RePro-generated rewritten instances are provided in Appendix~\ref{app:successful_cases}.

\subsection{Feasibility Screening}

Feasibility Screening begins with LLM-based rewriting and subsequently filters invalid candidate rewrites before executable formalization and proof verification. Given an original problem, the LLM generates a candidate rewrite while preserving its core mathematical structure, solution logic, difficulty, and answer type. The rewriting process applies strategies including numerical reparameterization, logical restructuring, constraint modification, and contextual reconstruction for word problems. The rewriter is instructed not to solve the rewritten problem or generate its new answer.

The generated candidate is then screened for problem validity. This step removes questions that are ill-defined, ambiguous, internally inconsistent, contradictory, unrealistic, or infeasible under basic real-world or task-specific constraints. Here, feasibility refers to the semantic and constraint consistency of the rewritten problem, rather than merely whether a target can be formally derived from its premises. Accordingly, candidates with contradictory premises are rejected regardless of whether the target is formally derivable from them. Detailed rewriting prompts, screening criteria, and implementation details are provided in Appendix \ref{app:prompt_templates}.

For example, in problems involving population counts or quantity constraints, automatic rewriting may introduce negative values or other conditions that violate implicit real-world assumptions \cite{zhou2026engiagent}. Although such outputs may appear mathematically expressible, they are considered infeasible under the intended problem semantics and are therefore removed. This stage is intended to ensure problem validity rather than answer correctness. Answer correctness is established later through Proof-level Verification, while the final answer-target matching step provides an additional target-level consistency check by ensuring that the verified answer corresponds to the specific quantity requested in the rewritten problem.

\subsection{Executable Formalization}

Executable Formalization converts rewritten instances that pass feasibility screening into machine-verifiable formal statements in Lean. This stage consists of a format check and a semantic check. The format check ensures that the generated Lean code is syntactically valid and can be compiled in Lean. The semantic check performs an LLM-assisted conservative alignment screen between the rewritten natural-language problem and the Lean formal statement, comparing key elements such as quantities, conditions, logical structure, object type, and the requested target.

We adopt a conservative all-pass policy. For each compiled formalization, the semantic checker is queried three times with temperature zero. A candidate is accepted only when all three judgments return Consistent. Any mismatch, parsing failure, or uncertain output causes the candidate to be discarded and regenerated.

This step should not be interpreted as formal verification of natural-language-to-Lean equivalence. It serves only as a pre-proof filter. RePro mitigates this limitation through subsequent proof-level verification and final-answer alignment: the ATP-generated proof must pass Lean checking, and the extracted answer must match the target quantity requested by the rewritten problem. Therefore, answer correctness is established only after Lean verification and answer alignment, while semantic consistency is conservatively screened rather than formally guaranteed. More details on the semantic screening policy and proof-grounded target-answer matching are provided in Appendices \ref{app:semantic_screening} and \ref{sec: Obtaining the Final Answer from Verified Proofs}.

\subsection{Proof-level Verification}

Proof-level Verification validates the correctness of reference answers through formal proof verification. This stage takes executable formalizations as input and retains only instances whose answers can be successfully proven and verified in Lean, forming the final evaluation dataset. Unlike earlier stages, Proof-level Verification is the only stage that determines answer correctness.

At this stage, ATP is used to construct candidate proofs for the formalized problems, which are then verified in Lean. Only instances whose answers can be verified by a valid proof are retained, while those that fail proof verification are discarded and regenerated. All correctness guarantees for reference answers originate from this stage.

As illustrated in Fig.~\ref{fig: Framework}, Operation~8 performs a constrained answer-alignment step between the rewritten problem and the valid proof. It extracts a candidate answer from spans that already appear in the proof and checks whether it matches the quantity requested in the problem. No additional computation, normalization, simplification, or inference is allowed at this stage. Only answers that exactly match the requested quantity are accepted; intermediate values, answers to a different target, and unresolved cases are discarded. This step is used solely to ensure target-answer consistency. Detailed descriptions are provided in Appendix~\ref{sec: Obtaining the Final Answer from Verified Proofs}.

\section{Evaluation Criteria}\label{sec: Evaluation Metrics}

\begin{figure*}[]
\centering
\includegraphics[width=1\textwidth]{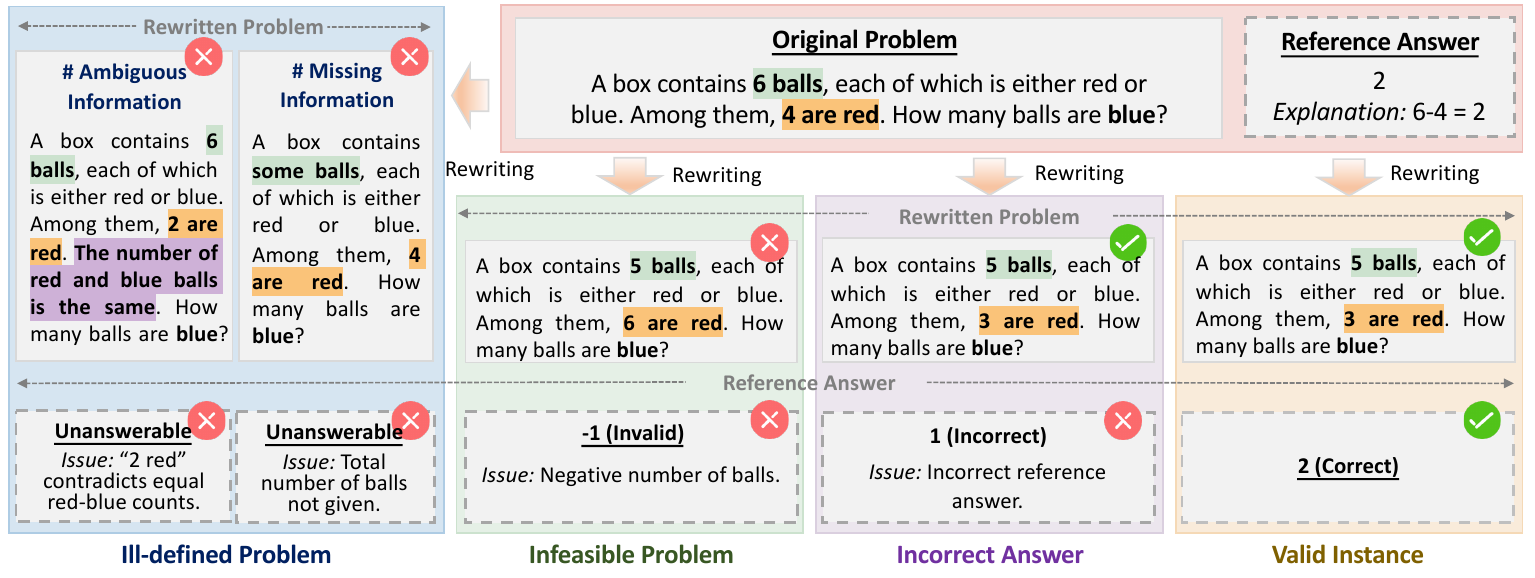}  %
\caption{Failure modes of rewriting-based evaluation and our solution. Rewriting may introduce three types of reliability issues: (1) ill-defined problems caused by ambiguous or missing information, (2) infeasible problems due to conflicting constraints, and (3) incorrect answers where the reference answer is wrong. A rewritten instance is valid only if the problem is well-defined, feasible, and paired with a correct reference answer.
}
\label{fig: failure_modes}
\end{figure*}

A benchmark problem can serve as a reliable evaluation instance only if the problem itself is valid and its reference answer is correct. The former requires the problem to be clearly specified, logically coherent, and solvable, while the latter ensures that model outputs are compared against a correct ground-truth answer. This is especially important for rewritten problems, where rewriting may alter not only surface wording but also problem semantics, constraints, or answer consistency. Recent benchmark verification studies further show that many evaluation failures arise from ambiguous statements, missing information, or incorrect reference answers.

Motivated by these observations, we evaluate rewritten problems from two complementary perspectives: \textit{problem validity} and \textit{answer correctness}. Problem validity includes two criteria: \textit{well-definedness}, requiring the problem to be clear and complete, and \textit{feasibility}, requiring it to admit a solution under the given conditions and satisfy basic real-world or task-specific constraints. Answer correctness requires that the reference answer be correct. Together, these define three criteria for a reliable benchmark instance: well-definedness, feasibility, and answer correctness. Fig.~\ref{fig: failure_modes} illustrates the corresponding failure modes and a valid rewritten instance. We therefore use three criteria:

\noindent \textbf{Well-definedness.} Measures whether the problem statement provides sufficient and unambiguous information to determine the task and its objective. A problem is considered not well-defined if it contains missing conditions, semantic ambiguity, unclear objects or variables, incomplete constraints, or an unspecified solving target.

\noindent \textbf{Feasibility.} Measures whether a well-defined problem admits a valid solution under basic real-world or task-specific constraints. A problem is considered infeasible if no valid solution exists or if the derived result violates these constraints due to conflicting conditions or inconsistencies.

\noindent \textbf{Answer Correctness.}
Measures whether the reference answer is formally verified. 
A reference answer is correct only if the candidate is successfully formalized, verified by a valid proof, and matched to the target quantity. 
In RePro, all retained candidates satisfy this requirement. 
In more general settings, candidates that fail automatic formalization should receive human-assisted checking to avoid hallucinated or unverifiable instances.

Let $N$ denote the total number of generated rewritten instances, and $N_c$ the number of instances satisfying criterion $c \in \{\text{well-definedness}, \text{feasibility}, \text{answer correctness}\}$. 
The corresponding rate is computed as $N_c / N$. To assess these criteria, we use a verification pipeline based on Lean, ATP, and LLM screening, with human assistance for ambiguous cases.

\begin{table*}[t]
\centering
\small
\setlength{\tabcolsep}{4pt}
\renewcommand{\arraystretch}{0.95}

\caption{Comparison of rewriting quality across methods on GSM8K and MATH. 
Metrics include well-definedness (Well-defined), feasibility (Feasible), answer correctness (Correct), and generation rate (Gen. Rate). 
The upper table reports results on the full set of generated rewritten instances, while the lower table reports results on the subset of benchmark instances for which RePro successfully generates rewritten problems.}
\label{tab:rewrite_quality}

\begin{tabular}{
l
>{\columncolor{gray!15}}c
>{\columncolor{gray!15}}c
>{\columncolor{gray!15}}c
>{\columncolor{gray!15}}c
c c c c
}

\toprule
\multicolumn{9}{c}{\textbf{Full Dataset}} \\

\toprule
\multirow{2}{*}{Method} 
& \multicolumn{4}{c}{GSM8K} 
& \multicolumn{4}{c}{MATH} \\

\cmidrule(lr){2-5} \cmidrule(lr){6-9}

& Well-defined $\uparrow$ 
& Feasible $\uparrow$ 
& Correct $\uparrow$ 
& Gen. Rate $\uparrow$
& Well-defined $\uparrow$ 
& Feasible $\uparrow$ 
& Correct $\uparrow$ 
& Gen. Rate $\uparrow$ \\

\midrule

Auto-Dataset  
& 99.60 & 99.19 & 87.10 & \textbf{100.00}
& 98.43 & 96.23 & 79.99 & \textbf{100.00} \\

ITD           
& \textbf{100.00} & 99.19 & 89.11 & \textbf{100.00}
& 98.26 & 97.85 & 81.15 & \textbf{100.00} \\

VarBench      
& 97.98 & 96.76 & 95.14 & 99.60
& 96.54 & 93.58 & 87.36 & 58.76 \\

\textbf{RePro (Ours)} 
& \textbf{100.00} & \textbf{100.00} & \textbf{100.00} & 88.31
& \textbf{100.00} & \textbf{100.00} & \textbf{100.00} & 59.16 \\

\bottomrule
\end{tabular}

\vspace{6pt}

\textbf{RePro Successful Generation Subset}

\vspace{2pt}

\begin{tabular}{
l
>{\columncolor{gray!15}}c
>{\columncolor{gray!15}}c
>{\columncolor{gray!15}}c
c
>{\columncolor{gray!15}}c
>{\columncolor{gray!15}}c
>{\columncolor{gray!15}}c
c
}

\toprule
\multirow{2}{*}{Method} 
& \multicolumn{4}{c}{GSM8K} 
& \multicolumn{4}{c}{MATH} \\

\cmidrule(lr){2-5} \cmidrule(lr){6-9}

& Well-defined $\uparrow$ 
& Feasible $\uparrow$ 
& Correct $\uparrow$ 
& Gen. Rate $\uparrow$
& Well-defined $\uparrow$ 
& Feasible $\uparrow$ 
& Correct $\uparrow$ 
& Gen. Rate $\uparrow$ \\

\midrule

Auto-Dataset  & 98.39 & 97.18 & 80.65 & \textbf{100.00} & 99.53 & 99.53 & 86.88 & \textbf{100.00} \\
ITD           & 97.58 & 96.77 & 81.05 & \textbf{100.00} & 100.00 & 99.07 & 87.66 & \textbf{100.00} \\

VarBench    &  98.79  & 98.39 &  96.77 & \textbf{100.00} & 95.49 & 93.73 & 89.23 & 60.53
 \\

\textbf{RePro (Ours)} 
& \textbf{100.00} & \textbf{100.00} & \textbf{100.00} & \textbf{100.00}
& \textbf{100.00} & \textbf{100.00} & \textbf{100.00} & \textbf{100.00} \\

\bottomrule
\end{tabular}
\end{table*}

\section{Experimental Methodology}

\subsection{Datasets}

To evaluate RePro, we select GSM8K \cite{cobbe2021gsm8k} and MATH \cite{hendrycks2021measuring} as benchmarks. We also considered harder benchmarks such as AIME \cite{dekoninck2026beyond} and Omni-MATH \cite{gao2025omni}, but current ATP bottlenecks led to very low generation rates; details are provided in Sec.~\ref{sec: Generation Rate Analysis} and Appendix~\ref{app:data_distribution}. 
Therefore, we focus on GSM8K and MATH, which are widely used for mathematical reasoning. 
GSM8K provides structured grade-school math problems, while MATH covers five Art of Problem Solving (AoPS) difficulty levels (LV1–LV5), ranging from basic high-school exercises to olympiad-level problems. Additional details on dataset selection and the data distribution before and after RePro filtering are provided in Appendix~\ref{app:data_distribution}. The metadata and fields of the released RePro dataset are described in Appendix~\ref{app:metadata}.

\subsection{Models}

We evaluate RePro on a diverse set of open-source language models from several major model families.  Specifically, we include Qwen2–0.5B/1.5B, Qwen3–0.6B/1.7B/8B/14B \cite{yang2025qwen3}, Llama-3.2–1B/3B \cite{touvron2023llama}, DeepSeek-R1–1.5B/7B/14B \cite{guo2025deepseek}, and Gemma 3–1B/4B \cite{team2025gemma}. This selection allows us to compare contamination-related behaviors across different architectures and training methods while keeping model scale controlled.

For the RePro generation pipeline, we use Qwen3-MAX \cite{yang2025qwen3} for problem rewriting and feasibility screening, Goedel-Formalizer-V2-8B for executable formalization into Lean 4 statements, and Goedel-Prover-V2-8B for ATP-based proof generation \cite{lin2025goedel}. The generated proofs are then verified by Lean 4.

\subsection{Baselines}

We compare our method with several representative approaches for automatic benchmark rewriting, including Auto-Dataset \cite{ying2024automating}, ITD \cite{zhu-etal-2024-inference}, and VarBench \cite{qian-etal-2024-varbench}. These methods generate new evaluation instances to mitigate benchmark leakage while preserving the original task structure. Auto-Dataset generates semantically similar questions from existing problems. ITD performs semantic-level rewriting while preserving the underlying numerical relations and computation logic. VarBench extracts numerical variables and constructs parameterized problem templates with executable solution functions, enabling new instances through variable resampling.

\section{Results}

\subsection{Rewriting Quality Comparison}\label{sec: Rewriting Quality Comparison}

\begin{figure}[]
    \centering
    \includegraphics[width=0.48\textwidth]{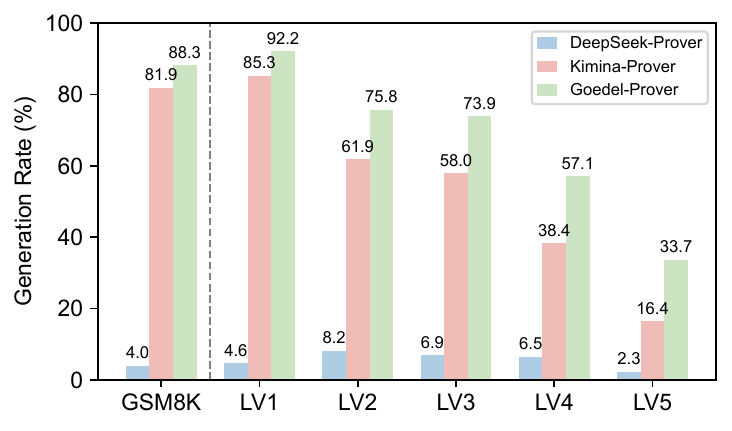}
    \caption{
    Impact of ATPs with varying capabilities on generation success rates. 
    }
    \label{fig:generation_rate_bar}
\end{figure}

\begin{figure*}[t]
    \centering
    \begin{minipage}{0.45\textwidth}
        \centering
        \includegraphics[width=\linewidth]{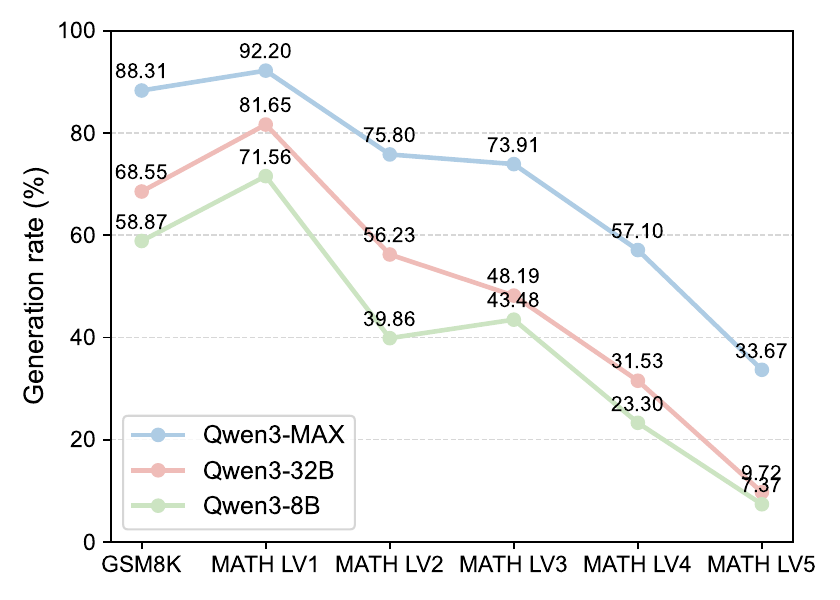}
        \caption{
        Effect of rewriter on generation rate.
        }
        \label{fig:rewriter_models}
    \end{minipage}
    \hfill
    \begin{minipage}{0.45\textwidth}
        \centering
        \includegraphics[width=\linewidth]{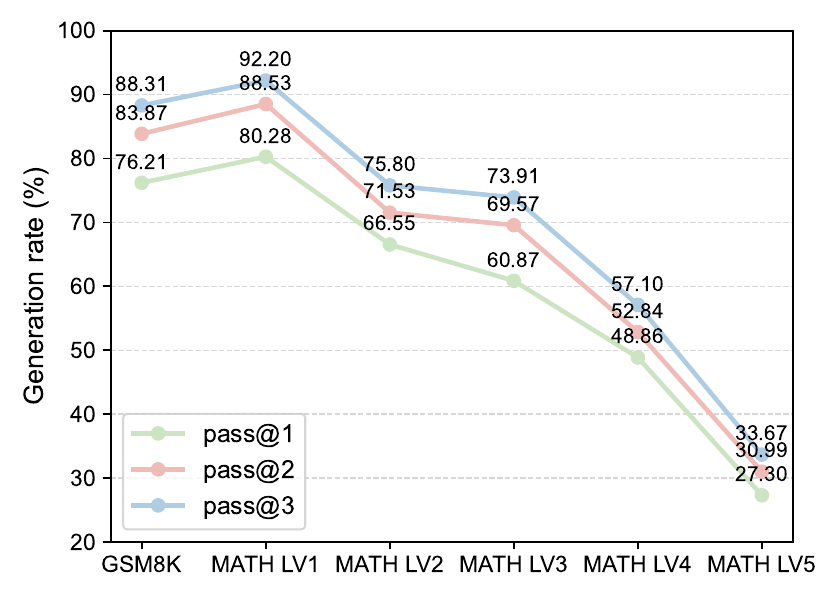}
        \caption{
        Effect of ATP call limit on generation rate.
        }
        \label{fig:repro_internal_passk}
    \end{minipage}
\end{figure*}

From Table~\ref{tab:rewrite_quality}, we obtain three key findings.

\noindent\textbf{RePro guarantees reliable retained instances.}
On both GSM8K and MATH, RePro achieves 100\% well-definedness, feasibility, and answer correctness. This shows that its verification pipeline removes ill-defined problems, infeasible instances, and incorrect reference answers from the retained rewritten set.

\noindent\textbf{Existing rewriting methods still produce invalid or incorrect instances.}
On the full datasets, AutoDataset, ITD, and VarBench achieve correctness rates of 79.99\%, 81.15\%, and 87.36\% on MATH, and 87.10\%, 89.11\%, and 95.14\% on GSM8K. Their feasibility rates also remain below 100\% on both datasets. These results show that existing methods can introduce incorrect references or invalid problems, undermining evaluation reliability. Representative candidate-level failure modes observed in our VarBench implementation are further analyzed in Appendix~\ref{app:varbench_failures}.

\noindent\textbf{RePro's gains are not due to subset selection alone.}
On the RePro-success subset, baselines still show non-trivial invalidity and incorrectness. Their correctness rates are 80.65\%, 81.05\%, and 96.77\% on GSM8K, and 86.88\%, 87.66\%, and 89.23\% on MATH, respectively. In contrast, RePro remains at 100\% across all criteria, indicating that the gains mainly come from its verification mechanism rather than subset selection.

As a supplementary reliability check, we independently audit 2,400 sampled rewritten instances across RePro and all baselines. Human judgments fully agree with the automatic reliability results. Details are provided in Appendix~\ref{app:human_validation}.

\subsection{Generation Rate Analysis} \label{sec: Generation Rate Analysis}

We analyze three key factors that affect RePro's generation rate: ATP capability, rewriter capability, and the ATP call limit. Overall, stronger ATPs and rewriters improve generation coverage, while increasing the ATP call limit brings additional but diminishing gains.

\subsubsection{Effect of ATP Capability}

RePro relies on ATPs to generate proofs for formalized problems, so ATP capability directly affects generation success. To study this effect, we keep all other components unchanged and only replace the ATP. We evaluate three strong ATPs below 15B parameters: DeepSeek-Prover-V2-7B (non-CoT) \cite{ren2025deepseek}, Kimina-Prover-Preview-Distill-7B \cite{wang2025kimina}, and Goedel-Prover-V2-8B \cite{lin2025goedel}, denoted as DeepSeek-Prover, Kimina-Prover, and Goedel-Prover, respectively. On miniF2F \cite{zheng2021minif2f}, their pass@32 success rates are 68.0\%, 63.1\%, and 84.6\%, respectively \cite{lin2025goedel,wang2025kimina}.

Fig.~\ref{fig:generation_rate_bar} shows that generation success rates generally increase with stronger ATP capability, while decreasing as problem difficulty increases from Level 1 to 5. We also observe that DeepSeek-Prover performs worse than expected given its miniF2F performance. Details are in Appendix~\ref{sec: Analysis of Failure Cases in DeepSeek-Prover-V2-7B}.

This result also highlights a quality-coverage trade-off. As shown in Table~\ref{tab:rewrite_quality}, VarBench achieves only 58.76\% generation rate on MATH due to stricter generation constraints. In contrast, RePro maintains a comparable generation rate of 59.16\% while ensuring both problem validity and answer correctness through formal verification. Overall, generation success is bounded by current ATP capability, and stronger ATPs are expected to further improve RePro's coverage.

\begin{figure*}[h]
    \centering
    \includegraphics[width=1\textwidth]{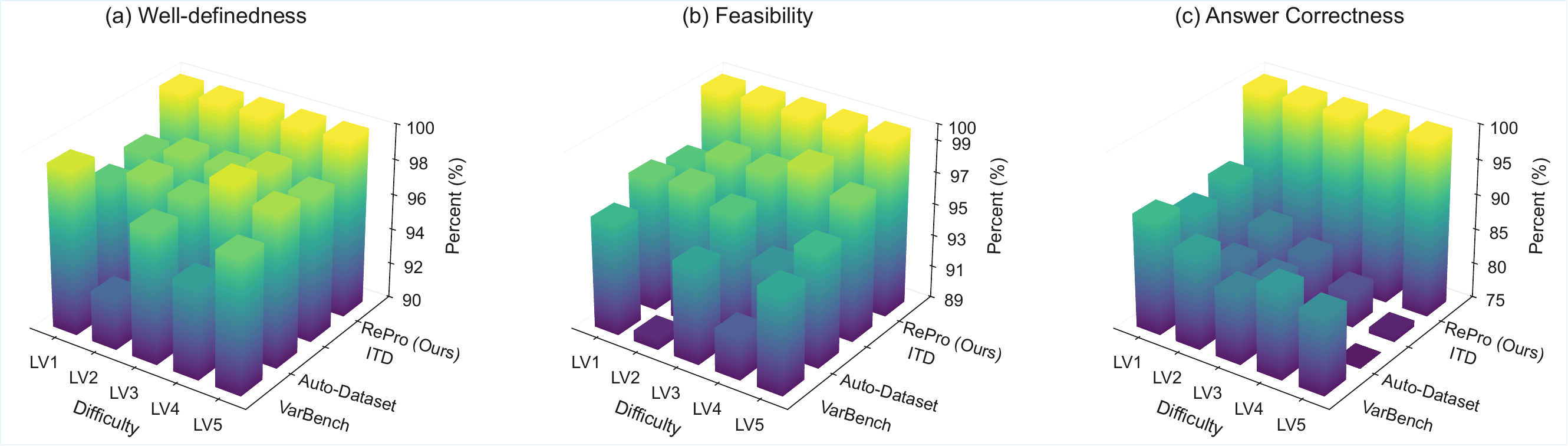}
    \caption{
    Rewriting quality comparison across five difficulty levels (LV1–LV5) of the MATH dataset. (a) Proportion of well-defined problems, (b) proportion of feasible problems, and (c) proportion of correct reference answers.
    }
    \label{fig:bar3d_metrics}
\end{figure*}

\begin{figure*}[]
    \centering
    \includegraphics[width=1\textwidth]{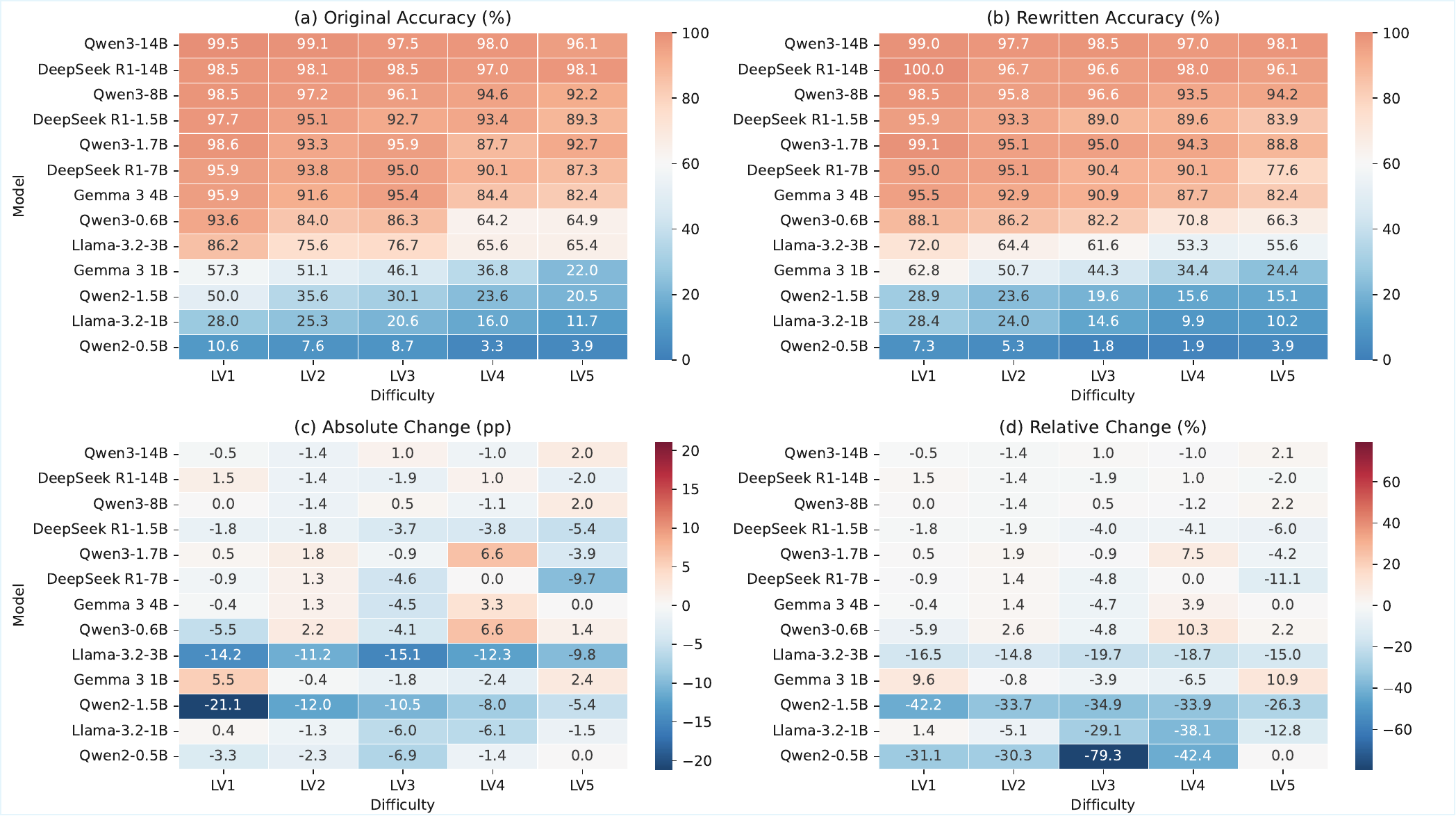}
    \caption{
    Impact of problem rewriting on model performance across five difficulty levels.
    (a) Original accuracy. (b) Accuracy after rewriting. (c) Absolute accuracy change (percentage points). (d) Relative accuracy change (percentage). Positive values indicate improvements, while negative values indicate performance drops.
    }
    \label{fig:rewriting_heatmap}
\end{figure*}

\subsubsection{Effect of Rewriter Capability}

We further analyze the effect of the rewriter model on RePro’s generation success rate. In this experiment, we keep the formalizer, ATP prover, Lean verification, and answer alignment settings unchanged, and only replace the model used for rewriting and feasibility screening.

As shown in Fig.~\ref{fig:rewriter_models}, the capability of the rewriter model has a clear impact on generation coverage. Qwen3-MAX achieves the highest success rate across all datasets and difficulty levels. For example, on GSM8K, Qwen3-MAX reaches 88.31\%, compared with 68.55\% for Qwen3-32B and 58.87\% for Qwen3-8B. A similar trend is observed across MATH difficulty levels, suggesting that stronger rewriters are more likely to produce candidates that can be successfully formalized and verified. Therefore, we use Qwen3-MAX as the default rewriter in the main experiments to obtain more stable generation coverage.

\subsubsection{Effect of ATP Call Limit}

We further examine how the ATP call limit affects RePro’s generation coverage. Here, pass@k allows up to k ATP proof-search attempts for each candidate that has passed rewriting, feasibility screening, and executable formalization. A sample is counted as successfully generated if at least one proof passes Lean verification and its verified answer matches the target answer.

As shown in Fig.~\ref{fig:repro_internal_passk}, increasing ATP calls consistently improves generation rate. On GSM8K, the rate increases from 76.21\% at pass@1 to 83.87\% at pass@2 and 88.31\% at pass@3. On MATH overall, it increases from 50.17\% to 55.49\% and 59.16\%, respectively. The same trend holds across all MATH difficulty levels, showing that additional ATP calls can recover part of the failures caused by unsuccessful proof generation.

The improvement from pass@2 to pass@3 is smaller than that from pass@1 to pass@2, suggesting diminishing returns as the ATP call limit increases. We therefore use pass@3 as the default setting in the main experiments to balance generation coverage and verification cost.

\subsection{Impact of Problem Difficulty on Rewriting Quality}

Fig.~\ref{fig:bar3d_metrics} shows how rewriting quality changes across MATH difficulty levels in terms of well-definedness, feasibility, and answer correctness. 

\noindent \textbf{Correctness decreases as problem difficulty increases.}
As shown in Fig.~\ref{fig:bar3d_metrics}(c), the correctness of ITD and AutoDataset drops noticeably as difficulty increases, from around 90\% at Level 1-2 to about 75\%–80\% at Level 4-5. This indicates that traditional rewriting methods are more likely to introduce incorrect reference answers as reasoning complexity grows.

\noindent \textbf{Well-definedness and feasibility remain stable, but invalid instances persist.}
As shown in Fig.~\ref{fig:bar3d_metrics}(a)(b), the well-defined and feasible rates of AutoDataset, ITD, and VarBench remain between 89\% and 100\% across all difficulty levels, without clear degradation as difficulty increases. However, ill-defined or infeasible problems still appear at every level, indicating that existing rewriting methods cannot fully eliminate invalid instances.

\subsection{Rewriting Sensitivity and Potential Memorization}

Fig.~\ref{fig:rewriting_heatmap} compares model accuracy on original MATH problems and their proof-verified rewritten counterparts. Since all retained RePro instances pass validity and answer-correctness verification, this paired comparison reduces the influence of invalid rewrites or incorrect reference answers. Thus, RePro provides a reliable diagnostic tool for analyzing model sensitivity to benchmark reformulation.

The results show that many models, especially smaller ones, lose accuracy after rewriting. For example, Qwen2-1.5B drops by 21.1, 12.0, and 10.5 percentage points on Level 1-3, respectively, while Llama-3.2-3B drops across all levels, with a maximum drop of 15.1 points. Meanwhile, some models improve after rewriting, often because rewritten problems clarify conditions, standardize notation, or reduce diagram-dependent difficulty. These mixed effects suggest that proof-verified rewriting reveals model-specific reformulation sensitivity.  Appendix~\ref{app:rewriting_sensitivity} analyzes potential confounds, showing near-zero correlations between surface or numeric changes and accuracy drop, and only weak positive correlations for solution and proof complexity.

\section{Conclusion}

In this work, we study the reliability of rewriting-based evaluation for contamination-resistant LLM benchmarking. Existing rewriting methods can reduce memorization effects, but often fail to ensure problem validity and answer correctness. We propose RePro, which integrates ATPs and proof-assistant checking into the rewriting pipeline. RePro retains only instances that can be successfully formalized and verified, ensuring problem validity and formally supported reference answers. Experiments on MATH and GSM8K show that RePro achieves 100\% well-defined, feasible, and correct retained instances, while existing methods still produce invalid problems or incorrect answers. Further analysis shows that proof-verified rewriting can reveal model-specific sensitivity to benchmark reformulation and provide signals of potential memorization or benchmark-specific pattern reliance.

\section*{Limitations}

While RePro provides the first framework that integrates automated theorem proving into benchmark rewriting and constructs evaluation datasets with proof-verified reference answers, several limitations remain that we plan to address in future work.

\noindent \textbf{Dependence on ATP capability.}
RePro relies on ATPs to verify candidate solutions during the rewriting. When the underlying ATP fails to find a valid proof, even correct and solvable problems may be filtered out, which can reduce the overall generation success rate. This limitation mainly reflects the current capability of neural theorem provers rather than the framework itself. As stronger ATP models continue to emerge, the generation coverage of RePro is expected to improve.

\noindent \textbf{Formalization constraints.}
RePro requires rewritten problems to be expressible in the Lean formal language in order to perform proof verification. As a result, the current framework mainly applies to problems that can be rewritten into a Lean representation. Tasks that rely heavily on natural language semantics or cannot be reasonably formalized in Lean, such as certain text-based reasoning problems, fall outside the current scope. Nevertheless, ongoing progress in automated formalization and proof assistants is expected to expand the range of tasks that can be supported.

\section*{Acknowledgements}
This work was supported in part by the Ministry of Education and Science of Bulgaria (support for INSAIT, part of the Bulgarian National Roadmap for Research Infrastructure), the Shenzhen Institute of Artificial Intelligence and Robotics for Society (AIRS), the Shenzhen Key Laboratory of Crowd Intelligence Empowered Low-Carbon Energy Network (No.~ZDSYS20220606100601002), the National Natural Science Foundation of China (No.~72331009).

\bibliography{custom}

@article{cheng2025survey,
  title={A survey on data contamination for large language models},
  author={Cheng, Yuxing and Chang, Yi and Wu, Yuan},
  journal={arXiv preprint arXiv:2502.14425},
  year={2025}
}

@inproceedings{chen-etal-2025-benchmarking-large,
    title = "Benchmarking Large Language Models Under Data Contamination: A Survey from Static to Dynamic Evaluation",
    author = "Chen, Simin  and
      Chen, Yiming  and
      Li, Zexin  and
      Jiang, Yifan  and
      Wan, Zhongwei  and
      He, Yixin  and
      Ran, Dezhi  and
      Gu, Tianle  and
      Li, Haizhou  and
      Xie, Tao  and
      Ray, Baishakhi",
    editor = "Christodoulopoulos, Christos  and
      Chakraborty, Tanmoy  and
      Rose, Carolyn  and
      Peng, Violet",
    booktitle = "Proceedings of the 2025 Conference on Empirical Methods in Natural Language Processing",
    month = nov,
    year = "2025",
    address = "Suzhou, China",
    publisher = "Association for Computational Linguistics",
    url = "https://aclanthology.org/2025.emnlp-main.511/",
    doi = "10.18653/v1/2025.emnlp-main.511",
    pages = "10080--10098",
    ISBN = "979-8-89176-332-6"
}

@inproceedings{li-etal-2024-open-source,
    title = "An Open-Source Data Contamination Report for Large Language Models",
    author = "Li, Yucheng  and
      Guo, Yunhao  and
      Guerin, Frank  and
      Lin, Chenghua",
    editor = "Al-Onaizan, Yaser  and
      Bansal, Mohit  and
      Chen, Yun-Nung",
    booktitle = "Findings of the Association for Computational Linguistics: EMNLP 2024",
    month = nov,
    year = "2024",
    address = "Miami, Florida, USA",
    publisher = "Association for Computational Linguistics",
    url = "https://aclanthology.org/2024.findings-emnlp.30/",
    doi = "10.18653/v1/2024.findings-emnlp.30",
    pages = "528--541"
}

@inproceedings{zhao-etal-2025-multimodal,
    title = "Are Multimodal {LLM}s Robust Against Adversarial Perturbations? {R}o{MM}ath: A Systematic Evaluation on Multimodal Math Reasoning",
    author = "Zhao, Yilun  and
      Gan, Guo  and
      Wang, Chengye  and
      Zhao, Chen  and
      Cohan, Arman",
    editor = "Chiruzzo, Luis  and
      Ritter, Alan  and
      Wang, Lu",
    booktitle = "Proceedings of the 2025 Conference of the Nations of the Americas Chapter of the Association for Computational Linguistics: Human Language Technologies (Volume 1: Long Papers)",
    month = apr,
    year = "2025",
    address = "Albuquerque, New Mexico",
    publisher = "Association for Computational Linguistics",
    url = "https://aclanthology.org/2025.naacl-long.582/",
    doi = "10.18653/v1/2025.naacl-long.582",
    pages = "11653--11665",
    ISBN = "979-8-89176-189-6"
}

@inproceedings{zhou2026engibench,
  title={Engibench: A benchmark for evaluating large language models on engineering problem solving},
  author={Zhou, Xiyuan and Wang, Xinlei and He, Yirui and Zou, Ruixi and Wu, Yang and Cheng, Yuheng and Xie, Yulu and Liu, Wenxuan and Zhao, Huan and Xu, Yan and others},
  booktitle={Findings of the Association for Computational Linguistics: ACL 2026},
  pages={36308--36334},
  year={2026}
}

@article{
hendrycks2021measuring,
title={Measuring Mathematical Problem Solving With the {MATH} Dataset},
author={Hendrycks, Dan and Burns, Collin and Kadavath, Saurav and Arora, Akul and Basart, Steven and Tang, Eric and Song, Dawn and Steinhardt, Jacob},
journal={arXiv preprint arXiv:2103.03874},
year={2021}
}

@article{cobbe2021gsm8k,
  title={Training Verifiers to Solve Math Word Problems},
  author={Cobbe, Karl and Kosaraju, Vineet and Bavarian, Mohammad and Chen, Mark and Jun, Heewoo and Kaiser, Lukasz and Plappert, Matthias and Tworek, Jerry and Hilton, Jacob and Nakano, Reiichiro and Hesse, Christopher and Schulman, John},
  journal={arXiv preprint arXiv:2110.14168},
  year={2021}
}

@article{yang2025qwen3,
  title={Qwen3 technical report},
  author={Yang, An and Li, Anfeng and Yang, Baosong and Zhang, Beichen and Hui, Binyuan and Zheng, Bo and Yu, Bowen and Gao, Chang and Huang, Chengen and Lv, Chenxu and others},
  journal={arXiv preprint arXiv:2505.09388},
  year={2025}
}

@article{touvron2023llama,
  title={LLaMA: Open and Efficient Foundation Language Models},
  author={Touvron, Hugo and Lavril, Thibaut and Izacard, Gautier and Martinet, Xavier and Lachaux, Marie-Anne and Lacroix, Timoth{\'e}e and Rozi{\`e}re, Baptiste and Goyal, Naman and Hambro, Eric and Azhar, Faisal and others},
  journal={arXiv preprint arXiv:2302.13971},
  year={2023}
}

@article{guo2025deepseek,
  title={DeepSeek-R1 incentivizes reasoning in LLMs through reinforcement learning},
  author={Guo, Daya and Yang, Dejian and Zhang, Haowei and Song, Junxiao and Wang, Peiyi and Zhu, Qihao and Xu, Runxin and Zhang, Ruoyu and Ma, Shirong and Bi, Xiao and others},
  journal={Nature},
  volume={645},
  number={8081},
  pages={633--638},
  year={2025},
  publisher={Nature Publishing Group UK London}
}

@article{team2025gemma,
  title={Gemma 3 technical report},
  author={Team, Gemma and Kamath, Aishwarya and Ferret, Johan and Pathak, Shreya and Vieillard, Nino and Merhej, Ramona and Perrin, Sarah and Matejovicova, Tatiana and Ram{\'e}, Alexandre and Rivi{\`e}re, Morgane and others},
  journal={arXiv preprint arXiv:2503.19786},
  year={2025}
}

@inproceedings{qian-etal-2024-varbench,
    title = "{V}ar{B}ench: Robust Language Model Benchmarking Through Dynamic Variable Perturbation",
    author = "Qian, Kun  and
      Wan, Shunji  and
      Tang, Claudia  and
      Wang, Youzhi  and
      Zhang, Xuanming  and
      Chen, Maximillian  and
      Yu, Zhou",
    editor = "Al-Onaizan, Yaser  and
      Bansal, Mohit  and
      Chen, Yun-Nung",
    booktitle = "Findings of the Association for Computational Linguistics: EMNLP 2024",
    month = nov,
    year = "2024",
    address = "Miami, Florida, USA",
    publisher = "Association for Computational Linguistics",
    url = "https://aclanthology.org/2024.findings-emnlp.946/",
    doi = "10.18653/v1/2024.findings-emnlp.946",
    pages = "16131--16161"
}

@inproceedings{
ying2024automating,
title={Automating Dataset Updates Towards Reliable and Timely Evaluation of Large Language Models},
author={Jiahao Ying and Yixin Cao and Yushi Bai and Qianru Sun and Bo Wang and Wei Tang and Zhaojun Ding and Yizhe Yang and Xuanjing Huang and Shuicheng YAN},
booktitle={The Thirty-eight Conference on Neural Information Processing Systems Datasets and Benchmarks Track},
year={2024},
url={https://openreview.net/forum?id=EvEqYlQv8T}
}

@inproceedings{butt2024benchagents,
  title={Benchagents: Automated benchmark creation with agent interaction},
  author={Butt, Natasha and Chandrasekaran, Varun and Joshi, Neel and Nushi, Besmira and Balachandran, Vidhisha},
  booktitle={ICLR 2025 Workshop on Navigating and Addressing Data Problems for Foundation Models},
  year={2024}
}

@inproceedings{cao-etal-2024-structeval,
    title = "{S}truct{E}val: Deepen and Broaden Large Language Model Assessment via Structured Evaluation",
    author = "Cao, Boxi  and
      Ren, Mengjie  and
      Lin, Hongyu  and
      Han, Xianpei  and
      Zhang, Feng  and
      Zhan, Junfeng  and
      Sun, Le",
    editor = "Ku, Lun-Wei  and
      Martins, Andre  and
      Srikumar, Vivek",
    booktitle = "Findings of the Association for Computational Linguistics: ACL 2024",
    month = aug,
    year = "2024",
    address = "Bangkok, Thailand",
    publisher = "Association for Computational Linguistics",
    url = "https://aclanthology.org/2024.findings-acl.314/",
    doi = "10.18653/v1/2024.findings-acl.314",
    pages = "5300--5318"
}

@inproceedings{zhu-etal-2024-inference,
    title = "Inference-Time Decontamination: Reusing Leaked Benchmarks for Large Language Model Evaluation",
    author = "Zhu, Qin  and
      Cheng, Qinyuan  and
      Peng, Runyu  and
      Li, Xiaonan  and
      Peng, Ru  and
      Liu, Tengxiao  and
      Qiu, Xipeng  and
      Huang, Xuanjing",
    editor = "Al-Onaizan, Yaser  and
      Bansal, Mohit  and
      Chen, Yun-Nung",
    booktitle = "Findings of the Association for Computational Linguistics: EMNLP 2024",
    month = nov,
    year = "2024",
    address = "Miami, Florida, USA",
    publisher = "Association for Computational Linguistics",
    url = "https://aclanthology.org/2024.findings-emnlp.532/",
    doi = "10.18653/v1/2024.findings-emnlp.532",
    pages = "9113--9129"
}

@article{shao2024deepseekmath,
  title={Deepseekmath: Pushing the limits of mathematical reasoning in open language models},
  author={Shao, Zhihong and Wang, Peiyi and Zhu, Qihao and Xu, Runxin and Song, Junxiao and Bi, Xiao and Zhang, Haowei and Zhang, Mingchuan and Li, YK and others},
  journal={arXiv preprint arXiv:2402.03300},
  year={2024}
}

@inproceedings{ahn-etal-2024-large,
    title = "Large Language Models for Mathematical Reasoning: Progresses and Challenges",
    author = "Ahn, Janice  and
      Verma, Rishu  and
      Lou, Renze  and
      Liu, Di  and
      Zhang, Rui  and
      Yin, Wenpeng",
    editor = "Falk, Neele  and
      Papi, Sara  and
      Zhang, Mike",
    booktitle = "Proceedings of the 18th Conference of the European Chapter of the Association for Computational Linguistics: Student Research Workshop",
    month = mar,
    year = "2024",
    address = "St. Julian{'}s, Malta",
    publisher = "Association for Computational Linguistics",
    url = "https://aclanthology.org/2024.eacl-srw.17/",
    doi = "10.18653/v1/2024.eacl-srw.17",
    pages = "225--237"
}

@inproceedings{wang-etal-2025-benchmark,
    title = "Benchmark Self-Evolving: A Multi-Agent Framework for Dynamic {LLM} Evaluation",
    author = "Wang, Siyuan  and
      Long, Zhuohan  and
      Fan, Zhihao  and
      Huang, Xuanjing  and
      Wei, Zhongyu",
    editor = "Rambow, Owen  and
      Wanner, Leo  and
      Apidianaki, Marianna  and
      Al-Khalifa, Hend  and
      Eugenio, Barbara Di  and
      Schockaert, Steven",
    booktitle = "Proceedings of the 31st International Conference on Computational Linguistics",
    month = jan,
    year = "2025",
    address = "Abu Dhabi, UAE",
    publisher = "Association for Computational Linguistics",
    url = "https://aclanthology.org/2025.coling-main.223/",
    pages = "3310--3328"
}

@inproceedings{rein2024gpqa,
  title={Gpqa: A graduate-level google-proof q\&a benchmark},
  author={Rein, David and Hou, Betty Li and Stickland, Asa Cooper and Petty, Jackson and Pang, Richard Yuanzhe and Dirani, Julien and Michael, Julian and Bowman, Samuel R},
  booktitle={First conference on language modeling},
  year={2024}
}

@article{phan2025lastexam,
      title = {A benchmark of expert-level academic questions to assess {AI} capabilities},
      author = {{Center for AI Safety} and {Scale AI} and {HLE Contributors Consortium}},
      journal = {Nature},
      volume = {649},
      pages = {1139--1146},
      year = {2026},
      doi = {10.1038/s41586-025-09962-4},
      eprint = {2501.14249},
      archivePrefix = {arXiv},
      primaryClass = {cs.LG},
      url = {https://arxiv.org/abs/2501.14249}
}

@article{zhai2026hle,
  title={HLE-Verified: A Systematic Verification and Structured Revision of Humanity's Last Exam},
  author={Zhai, Weiqi and Wang, Zhihai and Wang, Jinghang and Yang, Boyu and Li, Xiaogang and Xu, Xander and Wang, Bohan and Wang, Peng and Wu, Xingzhe and Li, Anfeng and others},
  journal={arXiv preprint arXiv:2602.13964},
  year={2026}
}

@article{ren2025deepseek,
  title={Deepseek-prover-v2: Advancing formal mathematical reasoning via reinforcement learning for subgoal decomposition},
  author={Ren, ZZ and Shao, Zhihong and Song, Junxiao and Xin, Huajian and Wang, Haocheng and Zhao, Wanjia and Zhang, Liyue and Fu, Zhe and Zhu, Qihao and Yang, Dejian and others},
  journal={arXiv preprint arXiv:2504.21801},
  year={2025}
}

@inproceedings{de2015lean,
  title={The Lean theorem prover (system description)},
  author={De Moura, Leonardo and Kong, Soonho and Avigad, Jeremy and Van Doorn, Floris and von Raumer, Jakob},
  booktitle={International Conference on Automated Deduction},
  pages={378--388},
  year={2015},
  organization={Springer}
}

@book{bertot2013interactive,
  title={Interactive theorem proving and program development: Coq’Art: the calculus of inductive constructions},
  author={Bertot, Yves and Cast{\'e}ran, Pierre},
  year={2013},
  publisher={Springer Science \& Business Media}
}

@article{wang2025kimina,
  title={Kimina-prover preview: Towards large formal reasoning models with reinforcement learning},
  author={Wang, Haiming and Unsal, Mert and Lin, Xiaohan and Baksys, Mantas and Liu, Junqi and Santos, Marco Dos and Sung, Flood and Vinyes, Marina and Ying, Zhenzhe and Zhu, Zekai and others},
  journal={arXiv preprint arXiv:2504.11354},
  year={2025}
}

@article{lin2025goedel,
  title={Goedel-prover-v2: Scaling formal theorem proving with scaffolded data synthesis and self-correction},
  author={Lin, Yong and Tang, Shange and Lyu, Bohan and Yang, Ziran and Chung, Jui-Hui and Zhao, Haoyu and Jiang, Lai and Geng, Yihan and Ge, Jiawei and Sun, Jingruo and others},
  booktitle={International Conference on Learning Representations},
  volume={2026},
  pages={11793--11818},
  year={2026}
}

@article{yang2023leandojo,
  title={Leandojo: Theorem proving with retrieval-augmented language models},
  author={Yang, Kaiyu and Swope, Aidan and Gu, Alex and Chalamala, Rahul and Song, Peiyang and Yu, Shixing and Godil, Saad and Prenger, Ryan J and Anandkumar, Animashree},
  journal={Advances in Neural Information Processing Systems},
  volume={36},
  pages={21573--21612},
  year={2023}
}

@article{zheng2021minif2f,
  title={Minif2f: a cross-system benchmark for formal olympiad-level mathematics},
  author={Zheng, Kunhao and Han, Jesse Michael and Polu, Stanislas},
  journal={arXiv preprint arXiv:2109.00110},
  year={2021}
}

@inproceedings{de2008z3,
  title={Z3: An efficient SMT solver},
  author={De Moura, Leonardo and Bj{\o}rner, Nikolaj},
  booktitle={International conference on Tools and Algorithms for the Construction and Analysis of Systems},
  pages={337--340},
  year={2008},
  organization={Springer}
}

@inproceedings{kovacs2013first,
  title={First-order theorem proving and Vampire},
  author={Kov{\'a}cs, Laura and Voronkov, Andrei},
  booktitle={International Conference on Computer Aided Verification},
  pages={1--35},
  year={2013},
  organization={Springer}
}

@article{lin2025zebralogic,
  title={Zebralogic: On the scaling limits of llms for logical reasoning},
  author={Lin, Bill Yuchen and Bras, Ronan Le and Richardson, Kyle and Sabharwal, Ashish and Poovendran, Radha and Clark, Peter and Choi, Yejin},
  journal={arXiv preprint arXiv:2502.01100},
  year={2025}
}

@article{zhu2025premise,
  title={Premise selection for a lean hammer},
  author={Zhu, Thomas and Clune, Joshua and Avigad, Jeremy and Jiang, Albert Qiaochu and Welleck, Sean},
  journal={arXiv preprint arXiv:2506.07477},
  year={2025}
}

@article{liu2023fimo,
  title={Fimo: A challenge formal dataset for automated theorem proving},
  author={Liu, Chengwu and Shen, Jianhao and Xin, Huajian and Liu, Zhengying and Yuan, Ye and Wang, Haiming and Ju, Wei and Zheng, Chuanyang and Yin, Yichun and Li, Lin and others},
  journal={arXiv preprint arXiv:2309.04295},
  year={2023}
}

@article{dekoninck2026beyond,
  title={Beyond Benchmarks: MathArena as an Evaluation Platform for Mathematics with LLMs},
  author={Dekoninck, Jasper and Jovanovi{\'c}, Nikola and Gehrunger, Tim and R{\"o}gnvalddson, K{\'a}ri and Petrov, Ivo and Sun, Chenhao and Vechev, Martin},
  journal={arXiv preprint arXiv:2605.00674},
  year={2026}
}

@inproceedings{gao2025omni,
  title={Omni-math: A universal olympiad level mathematic benchmark for large language models},
  author={Gao, Bofei and Song, Feifan and Yang, Zhe and Cai, Zefan and Miao, Yibo and Dong, Qingxiu and Li, Lei and Ma, Chenghao and Chen, Liang and Tang, Zhengyang and others},
  booktitle={International Conference on Learning Representations},
  volume={2025},
  pages={100540--100569},
  year={2025}
}

@article{spearman1904proof,
  title={The Proof and Measurement of Association between Two Things},
  author={Spearman, C},
  journal={The American Journal of Psychology},
  volume={15},
  number={1},
  pages={72--101},
  year={1904}
}

@article{wei2022chain,
  title={Chain-of-thought prompting elicits reasoning in large language models},
  author={Wei, Jason and Wang, Xuezhi and Schuurmans, Dale and Bosma, Maarten and Xia, Fei and Chi, Ed and Le, Quoc V and Zhou, Denny and others},
  journal={Advances in neural information processing systems},
  volume={35},
  pages={24824--24837},
  year={2022}
}

@inproceedings{zhou-etal-2024-paraphrase,
    title = "Paraphrase and Solve: Exploring and Exploiting the Impact of Surface Form on Mathematical Reasoning in Large Language Models",
    author = "Zhou, Yue  and
      Zhu, Yada  and
      Antognini, Diego  and
      Kim, Yoon  and
      Zhang, Yang",
    editor = "Duh, Kevin  and
      Gomez, Helena  and
      Bethard, Steven",
    booktitle = "Proceedings of the 2024 Conference of the North American Chapter of the Association for Computational Linguistics: Human Language Technologies (Volume 1: Long Papers)",
    month = jun,
    year = "2024",
    address = "Mexico City, Mexico",
    publisher = "Association for Computational Linguistics",
    url = "https://aclanthology.org/2024.naacl-long.153/",
    doi = "10.18653/v1/2024.naacl-long.153",
    pages = "2793--2804"
}

@inproceedings{yang-etal-2025-evaluating,
    title = "Evaluating Robustness of {LLM}s to Numerical Variations in Mathematical Reasoning",
    author = "Yang, Yuli  and
      Yamada, Hiroaki  and
      Tokunaga, Takenobu",
    editor = "Drozd, Aleksandr  and
      Sedoc, Jo{\~a}o  and
      Tafreshi, Shabnam  and
      Akula, Arjun  and
      Shu, Raphael",
    booktitle = "The Sixth Workshop on Insights from Negative Results in NLP",
    month = may,
    year = "2025",
    address = "Albuquerque, New Mexico",
    publisher = "Association for Computational Linguistics",
    url = "https://aclanthology.org/2025.insights-1.16/",
    doi = "10.18653/v1/2025.insights-1.16",
    pages = "171--180",
    ISBN = "979-8-89176-240-4"
}

@misc{wang2026memguardpersistingverifiersignals,
      title={MemGuard: Persisting Verifier Signals for LLM-Agent Memory Governance}, 
      author={Haoyu Wang and Guangyuan Dong and He Liang and Zijing Zhang and Jiachen Luo and Chuang Liu and Chao Xue and Hao Tang},
      year={2026},
      eprint={2608.21867},
      archivePrefix={arXiv},
      primaryClass={cs.AI},
      url={https://arxiv.org/abs/2608.21867}, 
}

@inproceedings{
zhou2026engiagent,
title={EngiAgent: Fully Connected Coordination of {LLM} Agents for Solving Open-ended Engineering Problems with Feasible Solutions},
author={Xiyuan Zhou and Ruixi Zou and Xinlei Wang and Yuheng Cheng and Yan Xu and Junhua Zhao and Jinjin Gu},
booktitle={Forty-third International Conference on Machine Learning},
year={2026},
url={https://openreview.net/forum?id=1p67QsYnbv}
}
\clearpage

\clearpage
\appendix

\section{The Use of Large Language Models}\

In this work, LLMs were used in four main ways:
\begin{enumerate}
    \item \textbf{Benchmark rewriting and feasibility screening.}
    LLMs were used to generate rewritten problem instances from existing benchmarks and to perform preliminary feasibility screening by identifying obviously invalid, ambiguous, or infeasible rewritten instances.

    \item \textbf{Semantic consistency checking during formalization.}
    LLMs were used to assess whether rewritten natural-language problems are semantically consistent with their corresponding Lean formal statements. This check serves only as a conservative pre-proof filter and does not provide formal verification of natural-language-to-Lean equivalence.

    \item \textbf{Answer extraction and target-answer matching.}
    LLMs were used to extract candidate final answers from Lean-verified proofs and check whether the extracted spans match the quantities requested by the rewritten problems. Candidate answers were restricted to spans that appear verbatim in the verified Lean proof, and no additional computation, simplification, or reasoning was allowed.

    \item \textbf{Figure and language assistance.}
    AI tools were used to generate some visual icons in Fig. \ref{fig: intro figure} and Fig. \ref{fig: Framework}, where applicable, other illustrative icons used in the figures. AI tools were also used for grammar checking and language refinement during the writing of this paper. All scientific claims, experimental results, and analyses were reviewed and verified by the authors.
\end{enumerate}

\section{Prompt Templates and Implementation Details}
\label{app:prompt_templates}

This appendix reports the main prompt templates and implementation details used in RePro. We include the prompts for problem rewriting, feasibility screening, semantic alignment screening, proof-grounded answer extraction, and final-answer classification. The templates are lightly formatted for readability. The released code contains the exact runtime prompts and parsing logic.

\paragraph{Problem rewriting prompt.}
The rewriting prompt asks the model to generate a new problem while preserving the underlying mathematical structure, solution logic, and answer type. It explicitly forbids solving the problem or computing the new answer.

\begin{lstlisting}[style=promptstyle]
You are a top-level exam question design expert. Your goal is to rewrite the given question while preserving its core mathematical structure, solution logic, and difficulty, but making it entirely new in form, wording, and surface meaning.

General requirements:
1. The rewritten question remains solvable and maintains a similar level of difficulty.
2. The underlying mathematical relationships, reasoning method, and solution logic remain equivalent.
3. The rewritten question is valid and logically consistent.
4. The rewritten question is entirely different from the original in surface form, wording, and semantics.
5. No part of the original phrasing, expressions, or narrative should be reused.
6. If a previous rewrite is provided, the new rewrite must be significantly different from it.

For word problems:
1. Completely change the story context or scenario.
2. Change the time, place, characters, identities, and physical quantities involved.
3. Use realistic and physically possible situations.
4. Choose a target unknown that is clearly well-defined and not semantically ambiguous.

For pure mathematical problems:
1. Keep the rewritten question purely mathematical.
2. Do not introduce a story or real-world context.

Rewriting strategies:
- Numerical reparameterization.
- Logical restructuring.
- Constraint modification.
- Contextual reconstruction for word problems.

Answer-type preservation:
The rewritten problem must preserve the type of the expected answer from the original problem, such as integer, rational number, interval, or finite set. Do not solve the problem or compute the answer. Enforce answer-type preservation structurally, for example by using linear expressions, proportional relationships, or factorizable polynomials when the original answer is rational or integral.

Return valid JSON only:
{
  "rewritten_question": "<the rewritten question as a string>"
}
\end{lstlisting}

The runtime user input is:

\begin{lstlisting}[style=promptstyle]
Original question:
<original problem>
\end{lstlisting}

\paragraph{Feasibility screening prompt.}
Feasibility screening is used as an early filter. Its purpose is to reject rewritten questions that are ill-defined, infeasible, ambiguous, internally inconsistent, or unrealistic. It does not establish reference-answer correctness; answer correctness is established later through Lean verification.

\begin{lstlisting}[style=promptstyle]
You are an expert in mathematics and careful problem interpretation.

Your task is not only to check whether the problem is mathematically solvable, but to judge whether it is well-defined, unambiguous, and valid under strict interpretation.

First classify the problem into one of two categories:

(A) Pure mathematical problem:
- No real-world story or physical interpretation is involved.
- Examples include solving equations, finding domains, simplifying expressions, algebraic manipulation, and function properties.

(B) Real-world or applied problem:
- The problem refers to people, objects, money, measurements, experiments, physical processes, or real-world actions.

Then apply the corresponding verification standard.

Your tasks:
1. Solve the problem.
2. Classify it as either (A) pure mathematical or (B) real-world/applied.
3. Judge whether the problem statement and the final numerical answer are valid under the appropriate strict standard.

For pure mathematical problems, check:
- Uniqueness.
- Mathematical clarity.
- Answer-type and format consistency.
- Mathematical reasonableness.

For real-world or applied problems, check:
- Uniqueness.
- Semantic clarity.
- Real-world executability.
- Unit and meaning consistency.
- Reasonableness.

For both categories:
If there is any ambiguity, vagueness, underspecification, incompatible condition, or mismatch between the mathematical answer and the required interpretation standard, the verdict must be "no".

Return valid JSON only:
{
  "analysis": "<solution and checks>",
  "verdict": "yes/no"
}
\end{lstlisting}

The runtime user input is:

\begin{lstlisting}[style=promptstyle]
Question:
<rewritten problem>
\end{lstlisting}

\paragraph{Formalization and proof-generation prompts.}
The formalizer and prover are called through backend models with short task instructions. The generated Lean statement must compile before semantic screening. A generated proof is accepted only if it passes Lean checking and does not contain \icode{sorry}.

\begin{lstlisting}[style=promptstyle]
Formalize the following question in Lean 4:
<rewritten problem>
\end{lstlisting}

\begin{lstlisting}[style=promptstyle]
Prove the following statement in Lean 4 without using 'sorry':
<Lean formal statement>
\end{lstlisting}

\paragraph{Semantic alignment screening prompt.}
The semantic alignment prompt checks whether a compiled Lean statement appears to encode the same task as the rewritten natural-language problem. This step is an LLM-assisted conservative screen rather than a formal proof of natural-language-to-Lean equivalence.

\begin{lstlisting}[style=promptstyle]
You are an expert in analyzing semantic consistency between a natural language math problem and its formal representation in Lean 4.

Your task is to judge whether the formal statement matches the natural language problem semantically.

You must not solve the problem, compute any values, or verify the correctness of the result. Your task is purely semantic.

A formalization is considered Consistent if and only if:
1. The same quantities are being referred to.
2. The same conditions are imposed.
3. The same logical structure is preserved.
4. The same type of object is being reasoned about, such as discrete vs continuous, individual vs aggregate, or instance vs range.
5. The same target quantity is being characterized.

Judge the formalization as Inconsistent if any of the following occurs:
A. Logical structure mismatch.
B. Wrong target quantity.
C. Quantitative mismatch in numbers, constraints, arithmetic relationships, units, or constants.
D. Missing condition.
E. Extra condition.

Special warning about rounding, ceiling, and floor:
If the natural language problem requires a rounding convention, such as "must buy whole items" or "round up", and the Lean formalization fails to encode that convention correctly, then it is inconsistent.

The natural-language meaning must match exactly. Even a small mismatch means Inconsistent.

Return valid JSON only:
{
  "judgment": "Consistent",
  "explanation": "<reason>"
}
or
{
  "judgment": "Inconsistent",
  "explanation": "<reason>"
}
\end{lstlisting}

The runtime user input is:

\begin{lstlisting}[style=promptstyle]
Natural language problem:
<rewritten problem>

Lean 4 formal statement:
<Lean theorem statement>
\end{lstlisting}

For each compiled Lean statement, the semantic checker is queried three times. The candidate is retained only when all three responses return \icode{Consistent}. Any \icode{Inconsistent} response, parsing failure, malformed output, or uncertain response leads to rejection and regeneration.

\paragraph{Proof-grounded answer extraction prompt.}
After a proof passes Lean verification, RePro extracts an answer candidate from the verified proof. The extractor is read-only: it may only copy spans that already appear in the Lean proof text.

\begin{lstlisting}[style=promptstyle]
You are a strict answer extractor for Lean 4 proof text.

You are given:
- The natural-language problem.
- The Lean 4 proof text.

Your job:
- Identify the exact final numeric answer or answers requested by the natural-language problem.
- You must not compute anything.
- You must not simplify anything.
- You must not evaluate arithmetic expressions.
- You must only copy answer candidates that literally appear in the Lean text.

If the natural-language problem asks for multiple quantities, then extract all of them and output them in one string separated by top-level commas.

Allowed:
- Copy existing expressions from the Lean text as-is.
- Select the expressions that match the quantity asked by the problem.

Not allowed:
- Arithmetic.
- Evaluation of powers or factorials.
- Symbolic simplification.
- Cancellation, expansion, or fraction reduction.
- Any inference requiring a new step.
- Using results from interactive commands such as #eval, #reduce, #check, or #print.

Return valid JSON only:
{
  "answer": "<string>",
  "evidence": "<snippet copied from Lean text>"
}

If no unique literal answer candidate can be found, or if the proof determines only part of the requested quantities, return:
{
  "answer": "unknown",
  "evidence": "unknown"
}
\end{lstlisting}

The runtime user input is:

\begin{lstlisting}[style=promptstyle]
Natural language problem:
<rewritten problem>

Lean 4 proof text:
<Lean proof>

Task:
Identify the requested quantity or quantities and extract the answer candidate only by copying from the Lean text. Do not compute or simplify. Do not use #eval, #reduce, #check, or #print results.
\end{lstlisting}

In addition to the prompt, RePro applies string-level checks to ensure that every comma-separated part of the extracted answer appears in the Lean proof text. Candidates that appear only near interactive commands such as \icode{\#eval}, \icode{\#reduce}, \icode{\#check}, or \icode{\#print} are rejected.

\paragraph{Final-answer classification prompt.}
Literal extraction alone is insufficient because a verified proof may contain intermediate values or values for a different target. RePro therefore uses a second classifier to determine whether the extracted candidate is the final answer requested by the rewritten problem.

\begin{lstlisting}[style=promptstyle]
You are a strict classifier for extracted answers from Lean 4 proof text.

You are given:
- The natural-language problem.
- The Lean proof text.
- An extracted answer candidate string that appears in the Lean text.

The candidate may contain multiple requested quantities separated by commas at top level.

Your job:
Do not compute or simplify anything. Decide whether the candidate is one of the following:

final:
Exactly the quantity or quantities asked, with no further computation, simplification, or evaluation needed.

intermediate:
An unfinished form that would require computation, simplification, or evaluation.

wrong_target:
A valid statement or value, but not the quantity requested by the problem.

unknown:
No unique final answer can be determined from the provided text or candidate, or the proof determines only part of the requested quantities.

Strict rules:
- Any candidate containing an unexecuted arithmetic operator or evaluation, such as +, -, *, /, ^, or !, should be treated as intermediate unless the problem explicitly asks for that exact expression form.
- Any candidate relying on #eval, #reduce, #check, or #print is unknown.
- If the problem asks for multiple quantities but the candidate provides fewer, classify it as unknown.
- Do not invent semantic conversions.
- Do not evaluate products or powers.

Return valid JSON only:
{
  "answer": "<string>",
  "status": "final|intermediate|wrong_target|unknown",
  "explanation": "<short reason>"
}
\end{lstlisting}

The runtime user input is:

\begin{lstlisting}[style=promptstyle]
Natural language problem:
<rewritten problem>

Lean 4 proof text:
<Lean proof>

Extracted candidate:
<candidate answer>

Evidence snippets:
<snippets copied from Lean text>

Now classify the candidate under the strict rules and output JSON only.
\end{lstlisting}

Only candidates classified as \icode{final} are accepted as reference answers. Candidates classified as \icode{intermediate}, \icode{wrong\_target}, or \icode{unknown} are rejected, and the corresponding rewritten instance is not retained.

\section{Semantic Alignment Screening and Target-Answer Matching}
\label{app:semantic_screening}

During executable formalization, RePro adopts a layered validation design to reduce the risk of semantic mismatch between the rewritten natural-language problem and the Lean formal statement. This design contains two complementary steps. First, before proof search, an LLM-assisted semantic alignment screen filters out Lean statements that are syntactically valid but appear semantically inconsistent with the rewritten problem. Second, after Lean verification, a proof-grounded target-answer matching step checks whether the extracted answer corresponds to the final quantity requested by the rewritten problem.

\paragraph{Semantic alignment screen.}
After a generated Lean statement passes compilation, RePro applies a conservative semantic alignment screen. The checker receives only the rewritten natural-language problem and the generated Lean 4 statement. It does not receive the original problem, the final answer, or the proof, and is explicitly instructed not to solve the problem, compute any value, or judge answer correctness. Instead, it checks whether the Lean statement preserves the same quantities, conditions, logical structure, object type, and target quantity as the rewritten problem.

A formalization is rejected if the checker detects mismatched constants, arithmetic relations, units, constraints, variable domains, logical connectives, target quantity, or required discrete operations such as rounding, ceiling, or floor behavior. Missing conditions and extra conditions are also treated as semantic mismatches.

\paragraph{Conservative all-pass policy.}
To reduce false acceptance, we adopt a conservative all-pass policy. For each compiled Lean statement, the semantic checker is queried three times using Qwen3-Max with temperature zero. A candidate passes semantic screening only if all three calls return \texttt{Consistent}. Any \texttt{Inconsistent} judgment, JSON parsing failure, malformed output, or uncertain response leads to rejection and regeneration. This policy makes the screen intentionally conservative: it may discard some valid formalizations, but it reduces the chance that an apparent semantic mismatch enters proof search.

\paragraph{Target-answer matching after Lean verification.}
Semantic screening checks whether the Lean statement appears to encode the same task as the rewritten problem, but it is not used to verify answer correctness. Therefore, RePro applies a separate target-answer matching step after proof-level verification. This step consists of two strictly constrained substeps: extracting an answer candidate from the Lean-verified proof and then checking whether the candidate is the final quantity requested by the rewritten problem.

During answer extraction, the candidate answer must be copied from literal spans that already appear in the verified proof. The extractor is not allowed to perform additional computation, normalization, simplification, or inference. Candidates that appear only in interactive commands such as \texttt{\#eval}, \texttt{\#reduce}, \texttt{\#check}, or \texttt{\#print} are rejected.

During answer classification, RePro determines whether the copied candidate is the final answer, an intermediate value, a wrong-target value, an unresolved expression, or unknown. A rewritten instance is retained only when the candidate is classified as \texttt{final} and matches the target quantity requested by the rewritten problem.

\paragraph{Reliability scope.}
The semantic alignment screen is not a formal proof of natural-language-to-Lean equivalence. Instead, it serves as a conservative pre-proof filter to reduce apparent semantic drift before formal verification. RePro does not rely on a single LLM judgment to establish retained-instance reliability. Instead, it uses multiple automated safeguards to reduce false acceptance: Lean compilation, conservative semantic screening, ATP-generated proof verification by Lean, proof-grounded answer extraction, and strict target-answer matching. The strongest formal guarantee applies to the Lean-verified proof for the accepted formal statement, while natural-language-to-Lean alignment and final answer-target alignment are supported by conservative LLM-assisted screening, proof-grounded extraction constraints, and strict classification.

\section{Obtaining the Final Answer from Verified Proofs}\label{sec: Obtaining the Final Answer from Verified Proofs}

\paragraph{Design goal.}
After a formally verified proof is obtained, the system must recover the final answer corresponding to the quantity requested in the rewritten problem.
A key requirement is that this post-proof stage must not introduce any new reasoning beyond the verified proof itself.

A naive design would directly generate the final answer from the proof.
However, this would allow post-hoc computation, simplification, or selection among intermediate results, effectively turning answer reporting into a second solving process.
Such behavior would weaken the RePro guarantee and reduce reproducibility.
To avoid this issue, post-proof answer recovery is divided into two constrained steps:
\emph{literal answer extraction} and \emph{final-answer checking}.

\paragraph{Literal answer extraction.}
The first step is purely read-only.
It extracts spans that appear verbatim in the verified proof and does not allow computation, normalization, inference, or rewriting.
Thus, the procedure may copy text from the proof, but may not derive new text.

For example, if the proof contains \texttt{have h : x = 17 := by ...},
then \texttt{17} can be extracted.
If the proof contains \texttt{2 + 3}, outputting \texttt{5} is not allowed.
If the proof contains \texttt{\{3\}}, outputting \texttt{3} is not allowed unless \texttt{3} also appears explicitly.

For problems with multiple target quantities,
all corresponding spans are returned, separated by commas,
without reordering or reformatting.

For instance, if the proof contains \texttt{a = 2} and \texttt{b = 5},
the extractor may output \texttt{2, 5},
but not \texttt{(2,5)} unless that exact form appears in the proof.

\paragraph{Final-answer checking.}
Literal extraction alone does not guarantee that the extracted span corresponds to the quantity requested in the problem,
since a verified proof may contain intermediate values,
auxiliary constants, or witness terms.

Therefore, a second step checks whether the extracted candidate matches the target quantity specified in the rewritten problem.
Importantly, this step is restricted to target alignment only:
it does not generate a new answer,
perform additional computation,
or replace the extracted span with a derived result.

For example, if the problem asks for \(x+y\)
and the proof contains \texttt{x = 3}, \texttt{y = 4}, and \texttt{x + y = 7},
then \texttt{3} and \texttt{4} are proof-grounded but not final,
while \texttt{7} is both proof-grounded and final.

If the problem asks for both \(a\) and \(b\),
and the proof contains \texttt{a = 2}, \texttt{b = 5}, and \texttt{a+b = 7},
then \texttt{2, 5} is final,
whereas \texttt{7} is not.

\paragraph{Summary.}
This design guarantees
(1) traceability to the verified proof,
(2) no answer generation after proof verification,
and (3) alignment with the quantity requested in the problem.

As a result, the final reported answer remains proof-grounded,
query-aligned, and free of post-hoc reasoning,
which preserves the RePro principle.

\section{Dataset Selection and Data Distribution}
\label{app:data_distribution}

\noindent
\textbf{Dataset selection.}
We select GSM8K and MATH for systematic evaluation because they remain widely used benchmarks for mathematical reasoning, contamination analysis, and benchmark rewriting, while also matching the current capability range of Lean-oriented neural ATPs. GSM8K contains structured grade-school math word problems with a relatively convergent solution space, making it suitable for controlled rewriting. MATH covers five AoPS difficulty levels, ranging from basic high-school exercises to olympiad-level problems, enabling evaluation across different reasoning complexities.

We also considered harder benchmarks such as AIME and Omni-MATH. However, \textbf{our preliminary experiment on 30 AIME 2025 problems \cite{dekoninck2026beyond} yields a RePro generation success rate of 0\%, suggesting that current ATP-based verification may still be insufficient for reliably constructing rewritten instances at this difficulty level}. This observation is also consistent with recent results on high-difficulty formal proof generation benchmarks such as FIMO \cite{liu2023fimo} and DeepSeek-ProverBench \cite{ren2025deepseek}, which are relevant to olympiad-style formal reasoning. Under pass@32, current 7B-8B Lean-oriented prover models achieve only 3.35-7.05\% on FIMO and 0.31-1.53\% on DeepSeek-ProverBench, as summarized in Table~\ref{tab:hard_benchmark_prover} \cite{lin2025goedel}. Therefore, we focus on the relatively more tractable GSM8K and MATH benchmarks in this work, and leave broader evaluation on harder benchmarks such as AIME and Omni-MATH to future work as ATP capabilities improve.

\begin{table}[h]
\centering
\small
\setlength{\tabcolsep}{3pt}
\begin{tabular}{lcc}
\toprule
Model & FIMO & DeepSeek-ProverBench \\
\midrule
DeepSeek-Prover-V2-7B & 5.70 & 0.31 \\
Kimina-Prover-7B & 3.35 & 1.38 \\
Goedel-Prover-V2-8B & 7.05 & 1.53 \\
\bottomrule
\end{tabular}
\caption{
Pass@32 success rates of recent Lean-oriented prover models on high-difficulty formal proof generation benchmarks. The low scores indicate a substantial ATP bottleneck for harder olympiad-style benchmarks.
}
\label{tab:hard_benchmark_prover}
\end{table}

\noindent
\textbf{Sampling protocol.}
Specifically, we adopt a rejection sampling procedure: candidate problems are randomly drawn from the source datasets and passed through the RePro pipeline. Only instances that satisfy all verification criteria are retained. This process continues until the number of valid rewritten instances reaches around 200 for each subset.

\begin{figure}[h]
    \centering
    \includegraphics[width=0.5\textwidth]{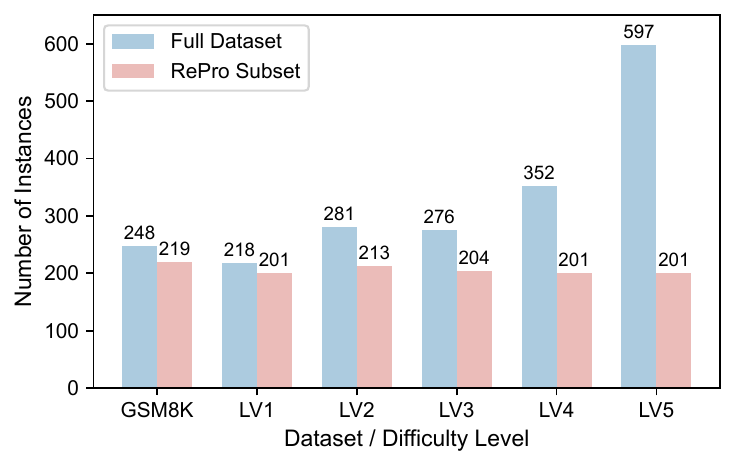}
    \caption{
Data distribution before and after RePro filtering.}
    \label{fig:Data distribution}
\end{figure}

\noindent
\textbf{Full dataset.}
The full dataset refers to all sampled candidate problems before filtering. As shown in Fig.~\ref{fig:Data distribution}, the number of sampled candidates varies across subsets because rejection sampling continues until approximately 200 verified instances are obtained for each subset. Consequently, harder levels such as LV4 and LV5 require substantially more sampled candidates due to their lower retention rates.

\noindent
\textbf{RePro successful generation subset.}
The final evaluation set consists only of instances that pass RePro verification. Due to the rejection sampling process, the resulting subset exhibits a nearly uniform distribution, with approximately 200 instances per difficulty level.

\noindent
\textbf{Retention-rate analysis.}
As shown in Fig.~\ref{fig:Data distribution}, the retention rate decreases substantially as problem difficulty increases. Specifically, RePro retains 88.3\% of sampled GSM8K candidates, while the retention rates on MATH are 92.2\%, 75.8\%, 73.9\%, 57.1\%, and 33.7\% for LV1--LV5, respectively. This pattern reveals a clear difficulty-dependent selection effect: under current formalization and proving capabilities, harder problems are less likely to pass the full verification pipeline. Consequently, although rejection sampling produces a nearly balanced final subset with approximately 200 instances per difficulty level, this retained subset does not preserve the original difficulty distribution of MATH. This observation further highlights that the current coverage of RePro is constrained by the capabilities of the underlying formalizer and ATP.

\noindent
\textbf{Reproducibility.}
All generated datasets, including rewritten problems and their formally verified proofs, are publicly released to facilitate reproducibility and further research.

\section{RePro Dataset Metadata}
\label{app:metadata}

Each instance in the RePro dataset corresponds to a rewritten problem derived from GSM8K or MATH, together with its reference answer and a formally verified proof. Only instances that pass the RePro verification pipeline are retained.

Each instance contains the following fields:

\begin{itemize}

\item \textbf{original\_question} – Original problem statement sampled from the source benchmark.

\item \textbf{original\_answer} – Ground-truth answer to the original problem.

\item \textbf{rewritten\_question} – Rewritten version of the original problem generated by the rewriting pipeline.

\item \textbf{new\_answer} – Reference answer corresponding to the rewritten problem.

\item \textbf{lean\_proof} – Formal proof in Lean that verifies the correctness of the rewritten problem and its answer.

\end{itemize}

\section{Representative Successful RePro Cases}
\label{app:successful_cases}

This section presents representative successful generation cases from RePro. Each case includes the original problem, the rewritten problem, the Lean formal statement, the verified answer, and a short explanation of why the instance passes the RePro verification pipeline. To avoid LaTeX compilation issues with Unicode symbols, we display Lean statements in ASCII form. The complete Lean proofs are included in the released dataset.

\paragraph{Case 1. GSM8K word problem with unit conversion.}

\textbf{Original problem.}
James buys 2 notebooks with 50 pages each. He pays \$5. How many cents did each page cost?

\textbf{Rewritten problem.}
A student purchases 4 sketchbooks, each containing 80 sheets of paper, for a total of \$16. What is the cost per sheet in cents?

\textbf{Verified answer.}
5

\textbf{Lean formal statement.}
\begin{lstlisting}[style=leanstyle]
theorem cost_per_sheet :
  let total_cost_cents : Rat := 16 * 100
  let total_sheets : Rat := 4 * 80
  total_cost_cents / total_sheets = 5 := by
\end{lstlisting}

\textbf{Verification outcome.}
The ATP-generated proof is accepted by Lean. The proof establishes that the total cost is 1600 cents, the total number of sheets is 320, and the unit cost is 5 cents per sheet.

\textbf{Why this instance passes RePro.}
The rewritten problem is well-defined because the requested quantity, cost per sheet in cents, is explicit. It is feasible because the monetary and counting quantities are realistic and internally consistent. The Lean statement preserves the same unit-conversion structure as the natural-language problem. The final answer is extracted from the verified proof and corresponds to the requested quantity.

\paragraph{Case 2. MATH inverse-function problem with a set-valued answer.}

\textbf{Original problem.}
Define \(f(x)=3x-8\). If \(f^{-1}\) is the inverse of \(f\), find the value or values of \(x\) for which \(f(x)=f^{-1}(x)\).

\textbf{Rewritten problem.}
Let \(g(x)=5x-12\). If \(g^{-1}\) denotes the inverse function of \(g\), determine all real numbers \(x\) such that \(g(x)=g^{-1}(x)\).

\textbf{Verified answer.}
\(\{3\}\)

\textbf{Lean formal statement.}
\begin{lstlisting}[style=leanstyle]
theorem g_inverse :
  let g : Real -> Real := fun x => 5 * x - 12
  let g_inv : Real -> Real := fun x => (x + 12) / 5
  {x : Real | g x = g_inv x} = ({3} : Set Real) := by
\end{lstlisting}

\textbf{Verification outcome.}
The ATP-generated proof is accepted by Lean. The proof verifies the equality between the solution set of \(g(x)=g^{-1}(x)\) and the singleton set \(\{3\}\).

\textbf{Why this instance passes RePro.}
The rewritten problem has a clear target: the full set of real solutions. The formal statement encodes the function, its inverse, and the requested solution set. The proof verifies set equality rather than merely showing that one candidate solution works. This ensures that the retained answer is proof-grounded and target-aligned.

\paragraph{Case 3. MATH symbolic factorization problem.}

\textbf{Original problem.}
Factor \(36-4x^2\) completely.

\textbf{Rewritten problem.}
Factor \(81-9y^2\) completely.

\textbf{Verified answer.}
\(9(3-y)(3+y)\)

\textbf{Lean formal statement.}
\begin{lstlisting}[style=leanstyle]
theorem factor_81_minus_9y2 (y : Real) :
  81 - 9 * y^2 = 9 * (3 - y) * (3 + y) := by
\end{lstlisting}

\textbf{Verification outcome.}
The ATP-generated proof is accepted by Lean. The proof verifies the algebraic identity between the original expression and the factored expression.

\textbf{Why this instance passes RePro.}
This case shows that RePro supports symbolic answers, not only numerical answers. The rewritten problem asks for a factored expression, and the Lean statement verifies the equivalence between the expanded and factored forms. The answer extraction step accepts the expression because the requested target is a symbolic factorization and the answer is grounded in the verified proof.

\paragraph{Case 4. MATH optimization problem with a minimum value.}

\textbf{Original problem.}
Square A and Square B are both \(2009\) by \(2009\) squares. Square A has both its length and width increased by an amount \(x\), while Square B has both its length and width decreased by the same amount \(x\). What is the minimum value of \(x\) such that the difference in area between the two new squares is at least as great as the area of a \(2009\) by \(2009\) square?

\textbf{Rewritten problem.}
Two identical square plots of land each measure \(1873\) meters on a side. One plot is expanded by adding \(y\) meters to both its length and width, while the other is reduced by subtracting \(y\) meters from both its length and width. What is the smallest positive value of \(y\) such that the absolute difference in area between the two modified plots is at least equal to the area of one original plot?

\textbf{Verified answer.}
\(1873/4\)

\textbf{Lean formal statement.}
\begin{lstlisting}[style=leanstyle]
theorem minimum_area_difference :
  let original_side : Real := 1873
  let expanded_area : Real -> Real :=
    fun y => (original_side + y)^2
  let reduced_area : Real -> Real :=
    fun y => (original_side - y)^2
  let original_area : Real := original_side^2
  let area_difference : Real -> Real :=
    fun y => abs (expanded_area y - reduced_area y)
  let condition : Real -> Prop :=
    fun y => And (y > 0) (original_area <= area_difference y)
  Exists (fun y : Real =>
    And (y = original_side / 4)
      (And (condition y)
        (forall z : Real, condition z -> y <= z))) := by
\end{lstlisting}

\textbf{Verification outcome.}
The ATP-generated proof is accepted by Lean. The proof verifies that \(y=1873/4\) satisfies the area-difference condition and that every positive value satisfying the condition is at least \(1873/4\).

\textbf{Why this instance passes RePro.}
This case demonstrates a more complex successful rewrite. The rewritten problem changes the numerical parameter and real-world context while preserving the optimization structure. The Lean statement encodes the positivity condition, the area inequality, and the minimality requirement. Therefore, the pipeline does not merely verify that the answer satisfies the inequality; it also verifies that it is the smallest valid value.

\paragraph{Summary.}
These cases illustrate different types of retained RePro instances. Case 1 shows a real-world arithmetic problem with unit conversion. Case 2 shows a set-valued algebraic answer. Case 3 shows a symbolic expression answer. Case 4 shows an optimization problem requiring a minimality proof. In each case, the rewritten problem passes feasibility screening, the Lean statement compiles, the semantic alignment screen accepts the formalization, the ATP-generated proof passes Lean verification, and the extracted answer corresponds to the target quantity requested by the rewritten problem.

\section{Reliability Interpretation and Human Validation}
\label{app:human_validation}

RePro is a verification-driven benchmark rewriting framework. Therefore, the reported 100\% well-definedness, feasibility, and answer correctness should be interpreted as reliability metrics for the retained rewritten instances, rather than as a claim that all raw LLM-generated rewrites are correct.

In RePro, generation and verification are explicitly decoupled. LLMs first generate candidate rewritten problems, and the verification pipeline then filters these candidates through feasibility screening, executable formalization, proof generation, Lean verification, and target-answer matching. A rewritten instance is retained only when its reference answer is supported by an ATP-generated proof that passes Lean kernel-level verification, and when the verified answer matches the target quantity requested in the rewritten problem. Therefore, the correctness metric measures whether the retained reference answers are backed by formally verified proof certificates.

This result should be interpreted together with the generation rate. Correctness measures the reliability of retained instances, while generation rate measures the coverage cost required to obtain such verified instances. In other words, RePro does not assume that all generated candidates are correct. Instead, it removes unreliable candidates and retains only those that satisfy the full verification pipeline.

To further examine whether the automatic verification toolchain introduces potential bias, we conduct an independent human validation for all rewriting methods. For each method, including RePro, Auto-Dataset, ITD, and VarBench, we randomly sample 600 rewritten instances, consisting of 100 instances from GSM8K and 100 instances from each difficulty level of MATH. In total, the human validation covers 2,400 rewritten instances.

For each sampled instance, human auditors are given only the original problem, the rewritten problem, and the reference answer. The automatic pipeline decisions, Lean formalizations, ATP outputs, and verified proofs are not used during human validation. Human auditors evaluate each instance according to the three reliability criteria defined in Sec.~\ref{sec: Evaluation Metrics}: well-definedness, feasibility, and answer correctness.

Table~\ref{tab:human_validation} reports the human validation results. The human judgments are fully consistent with the automatic pipeline judgments across all methods, datasets, and criteria. In particular, all sampled RePro-retained instances are confirmed to be well-defined, feasible, and paired with correct reference answers. For baseline methods, the human validation also confirms the invalid problems and incorrect reference answers identified by the automatic reliability evaluation. 

\begin{table*}[t]
\centering
\small
\setlength{\tabcolsep}{8pt}
\begin{tabular}{llrrrr}
\toprule
Method & Dataset & \#Checked 
& Well-defined Agreement 
& Feasibility Agreement 
& Correctness Agreement \\
\midrule
Auto-Dataset & GSM8K & 100 & 100.00 & 100.00 & 100.00 \\
Auto-Dataset & MATH  & 500 & 100.00 & 100.00 & 100.00 \\
ITD          & GSM8K & 100 & 100.00 & 100.00 & 100.00 \\
ITD          & MATH  & 500 & 100.00 & 100.00 & 100.00 \\
VarBench     & GSM8K & 100 & 100.00 & 100.00 & 100.00 \\
VarBench     & MATH  & 500 & 100.00 & 100.00 & 100.00 \\
RePro        & GSM8K & 100 & 100.00 & 100.00 & 100.00 \\
RePro        & MATH  & 500 & 100.00 & 100.00 & 100.00 \\
\midrule
Total        & All   & 2400 & 100.00 & 100.00 & 100.00 \\
\bottomrule
\end{tabular}
\caption{
Human-machine agreement in the independent validation. For each rewriting method, we randomly sample 600 instances, including 100 from GSM8K and 100 from each MATH difficulty level. Human auditors independently judge well-definedness, feasibility, and answer correctness without using automatic pipeline decisions, Lean formalizations, ATP outputs, or verified proofs. The table reports the percentage of sampled instances for which human judgments agree with the automatic pipeline judgments for each criterion.
}
\label{tab:human_validation}
\end{table*}

\section{Confound Analysis for Rewriting Sensitivity}
\label{app:rewriting_sensitivity}

This section further analyzes possible factors behind performance changes after proof-verified rewriting. The goal is not to prove data contamination, but to examine whether these changes can be explained by simpler rewriting-induced factors, such as problem length change, surface-form change, numeric changes, solution complexity, and proof complexity. Direct evidence of data contamination would require overlap analysis against model training data or external corpora, which is beyond the scope of this work.

\paragraph{Data and unit of analysis.}
We conduct the analysis on the RePro-retained MATH evaluation set. Each retained instance contains an original problem, a proof-verified rewritten problem, and a corresponding Lean proof. For instance \(i\) and model \(m\), we define correctness on the original and rewritten problems as
\[
c_{i,m}^{ori}, c_{i,m}^{rew} \in \{0,1\},
\]
where 1 denotes a correct answer and 0 denotes an incorrect answer. The accuracy drop for this model-instance pair is defined as
\[
d_{i,m}=c_{i,m}^{ori}-c_{i,m}^{rew}.
\]
Thus, \(d_{i,m}=1\) means that the model answers the original problem correctly but fails on the rewritten problem, while \(d_{i,m}=-1\) means the opposite. To analyze instance-level confounds, we compute the average drop across evaluated models:
\[
\bar{d}_i=\frac{1}{M}\sum_{m=1}^{M}d_{i,m}.
\]

\paragraph{Overall observation.}
The results show a bidirectional pattern. Many models exhibit accuracy drops after rewriting, while some models improve on certain rewritten instances. This indicates that proof-verified rewriting does not produce a one-directional effect. Instead, it changes the evaluation distribution in multiple ways and should be analyzed as model-specific sensitivity to benchmark reformulation.

\paragraph{Confound metrics.}
For each original--rewritten pair, we compute several potential confound metrics. Problem length change measures whether the rewritten problem becomes longer or shorter than the original problem. We tokenize each problem statement and define
\[
\Delta L_q(i)=\log\frac{L_q^{rew}(i)+1}{L_q^{ori}(i)+1},
\]
where \(L_q^{ori}(i)\) and \(L_q^{rew}(i)\) denote the token lengths of the original and rewritten problems. We also consider \(|\Delta L_q(i)|\), which measures the magnitude of length change regardless of direction.

Surface-form distance measures how much the rewritten problem differs from the original at the string level \cite{zhou-etal-2024-paraphrase}. We lowercase both problem statements, collapse whitespace, and compute
\[
D_{surf}(i)=1-\mathrm{sim}(q_i^{ori}, q_i^{rew}),
\]
where \(\mathrm{sim}\) is the normalized sequence-matching similarity score. Larger values indicate greater surface-form divergence.

Numeric-range change measures whether rewriting changes the numerical scale of the problem \cite{yang-etal-2025-evaluating}. Let \(A^{ori}(i)\) and \(A^{rew}(i)\) be the maximum absolute numeric values in the original and rewritten problem, respectively. We define
\[
\Delta A(i)=\log(1+A^{rew}(i))-\log(1+A^{ori}(i)).
\]
We also consider \(|\Delta A(i)|\). Numeric-count change measures whether rewriting introduces more or fewer numeric quantities:
\[
\Delta N(i)=N^{rew}(i)-N^{ori}(i),
\]
where \(N^{ori}(i)\) and \(N^{rew}(i)\) denote the numbers of numeric literals in the original and rewritten problems.

Solution-length change is used as a proxy for natural-language solution complexity \cite{wei2022chain}. Let \(L_s^{ori}(i)\) and \(L_s^{rew}(i)\) denote the token lengths of the original and rewritten solutions. We define
\[
\Delta L_s(i)=\log\frac{L_s^{rew}(i)+1}{L_s^{ori}(i)+1}.
\]
We also report the rewritten solution length:
\[
L_s^{rew,log}(i)=\log(1+L_s^{rew}(i)).
\]
Finally, Lean proof length is used as a lightweight proxy for formal proof complexity \cite{zheng2021minif2f}. Let \(L_p(i)\) be the token length of the verified Lean proof. We define
\[
L_p^{log}(i)=\log(1+L_p(i)).
\]

 
\paragraph{Correlation analysis.}
Since length, numeric values, and proof lengths can be heavy-tailed, we use Spearman correlation rather than Pearson correlation \cite{spearman1904proof}. For each confound metric, we compute its correlation with the model-averaged accuracy drop \(\bar{d}_i\). Table~\ref{tab:confound_corr} reports the results.

\begin{table}[t]
\centering
\small
\setlength{\tabcolsep}{9pt}
\begin{tabular}{lr}
\toprule
Confound metric & \makecell{Spearman $\rho$ \\ with accuracy drop} \\
\midrule
Problem length change & -0.000 \\
Absolute problem length change & -0.031 \\
Surface-form distance & -0.031 \\
Numeric-range change & 0.014 \\
Absolute numeric-range change & 0.002 \\
Numeric-count change & -0.021 \\
Absolute numeric-count change & -0.032 \\
Solution-length change & 0.101 \\
Absolute solution-length change & 0.099 \\
Lean proof length & 0.135 \\
Rewritten solution length & 0.173 \\
\bottomrule
\end{tabular}
\caption{
Spearman correlations between potential confound metrics and model-averaged accuracy drop on the RePro-retained MATH evaluation set. Surface-level and numeric changes have near-zero correlations with accuracy drop, while solution and proof complexity show weak positive correlations.
}
\label{tab:confound_corr}
\end{table}

The surface-level metrics have correlations close to zero. Problem length change, absolute problem length change, and surface-form distance are not meaningfully associated with accuracy drop. This suggests that the observed drops are not primarily explained by rewritten problems being longer or more surface-dissimilar.

The numeric metrics also have near-zero correlations. Numeric-range change, absolute numeric-range change, numeric-count change, and absolute numeric-count change are weakly associated with accuracy drop. This suggests that the observed drops are not mainly driven by larger numbers, changed numerical ranges, or increased numbers of numeric quantities.

In contrast, complexity-related metrics show weak positive correlations. Solution-length change, rewritten solution length, and Lean proof length are positively associated with accuracy drop. This indicates that some rewritten problems may become harder because they require longer solutions or more complex formal proofs. Therefore, solution and proof complexity remain plausible contributing factors.

\paragraph{Improvement cases.}
To better understand why some models achieve higher accuracy on the rewritten benchmark, we manually inspect representative improvement cases from Level 4, where models are more likely to answer correctly after rewriting. These examples suggest that such improvements are not necessarily caused by reduced mathematical difficulty. Instead, they often arise because rewriting makes the target quantity, condition structure, or information flow easier to parse.

\begin{tcolorbox}[failurebox,
  title=\exampletitle{Target Quantity Clarification},
  width=\linewidth,
  breakable]
\small
\textbf{Original problem.}
What value of $x$ will give the maximum value for $-x^2 - 6x + 12$?

\medskip
\textbf{Original answer:} $-3$

\medskip
\textbf{Rewritten problem.}
For what value of $x$ does the expression $-2x^2 + 16x - 5$ attain its maximum value?

\medskip
\textbf{Rewritten answer:} $4$

\medskip
\textbf{Interpretation.}
The original wording may confuse the target: some models may output the maximum function value instead of the \(x\)-value that gives the maximum. The rewrite states the target more directly. This reflects target-identification sensitivity, not lower mathematical difficulty.
\end{tcolorbox}

\begin{tcolorbox}[failurebox,
  title=\exampletitle{Explicit Condition Phrasing},
  width=\linewidth,
  breakable]
\small
\textbf{Original problem.}
For specific positive numbers $m$ and $n$, the quadratics $16x^2+36x+56$ and $(mx+n)^2$ differ only in their constant term. What is $mn$?

\medskip
\textbf{Original answer:} $18$

\medskip
\textbf{Rewritten problem.}
For certain positive integers $p$ and $q$, the quadratic expressions $81x^2+108x+64$ and $(px+q)^2$ have identical coefficients for $x^2$ and $x$, but their constant terms are not equal. Compute $pq$.

\medskip
\textbf{Rewritten answer:} $54$

\medskip
\textbf{Interpretation.}
Both problems require the same algebra step: expand the square, match the \(x^2\) and \(x\) coefficients, and compute the product. The original phrase ``differ only in their constant term'' requires the model to infer which coefficients should be matched. The rewrite states this directly by saying that the \(x^2\) and \(x\) coefficients are identical. This makes the condition clearer without making the algebra easier.
\end{tcolorbox}

\begin{tcolorbox}[failurebox,
  title=\exampletitle{Condition Tracking Clarification},
  width=\linewidth,
  breakable]
\small
\textbf{Original problem.}
Annie is located at $(3,5)$ and Barbara says she is located at $(-6,2)$. They agree to meet at the midpoint of their current locations. However, Barbara read the map wrong and is actually at $(-10,4)$. What is the positive difference in the $x$-coordinates of where they agreed to meet and where they should actually meet?

\medskip
\textbf{Original answer:} $2$

\medskip
\textbf{Rewritten problem.}
Annie is at $(-2,7)$, and Carlos initially reports his location as $(4,-1)$. Based on this, they decide on a meeting point. Later, Carlos realizes he is actually at $(6,-5)$. What is the absolute difference between the $x$-coordinates of the originally planned meeting point and the correct meeting point?

\medskip
\textbf{Rewritten answer:} $1$

\medskip
\textbf{Interpretation.}
Both problems require computing two midpoints and comparing their \(x\)-coordinates. In the original problem, the model must infer that ``where they agreed to meet'' means the midpoint based on the reported location, and ``where they should actually meet'' means the midpoint based on the corrected location. The rewrite names these two quantities directly as the ``originally planned meeting point'' and the ``correct meeting point'', making the target clearer.
\end{tcolorbox}

\paragraph{Interpretation.}
Overall, this analysis supports a cautious interpretation of the rewriting results. Simple problem length changes, surface-form differences, and numeric changes do not fully explain the observed performance drops. However, solution and proof complexity show weak positive associations with drops, meaning that increased reasoning complexity remains a plausible confound.

Therefore, performance changes after proof-verified rewriting are better interpreted as a reliability-controlled diagnostic of model sensitivity to benchmark reformulation. Accuracy drops may suggest potential memorization or benchmark-specific pattern reliance, while improvements may reflect reduced incidental difficulty or clearer problem specification. Direct claims about data contamination require additional overlap analysis against model training data or external corpora.

\section{Analysis of Failure Cases in DeepSeek-Prover-V2-7B}\label{sec: Analysis of Failure Cases in DeepSeek-Prover-V2-7B}

This section presents representative failure cases observed in the Lean
proofs generated by DeepSeek-Prover-V2-7B. These cases illustrate
several failure patterns that may contribute to the gap between the
prover's reported benchmark capability and its performance in our
rewriting pipeline.

\begin{tcolorbox}[failurebox,
  title=\exampletitle{Inappropriate Tactic Selection}]

The generated proof fails to select tactics appropriate for the current
goal. In particular, it repeatedly applies generic arithmetic tactics
that do not establish the required polynomial identity over the reals.

\smallskip
\textbf{Question}

Simplify the expression $(x+5)(x-4) - x(x-1)$.

\smallskip
\textbf{Failed Proof}

\begin{lstlisting}[style=leanstyle]
import Mathlib
import Aesop

set_option maxHeartbeats 200000

open scoped BigOperators
open BigOperators Real Nat Topology Rat

theorem simplify_expr (x : Real) :
  (x + 5) * (x - 4) - x * (x - 1) = 2 * x - 20 := by
  norm_num
  <;> omega
  <;> omega
  <;> omega
  <;> omega
  <;> omega
-- omitted (over 100 lines)
\end{lstlisting}

\end{tcolorbox}

In this example, the theorem statement is mathematically correct, but
the generated proof does not use an appropriate algebraic
normalization tactic such as \texttt{ring} or \texttt{ring\_nf}.
Instead, it repeatedly applies \texttt{norm\_num} and \texttt{omega},
which do not resolve the symbolic polynomial identity over
\texttt{Real}. We observe similar failure patterns in which the prover
falls back to generic tactics such as \texttt{norm\_num},
\texttt{omega}, \texttt{rfl}, or \texttt{decide} without making
goal-directed progress. These cases suggest that the prover does not
always adapt its tactic selection to the current proof state and may
instead revert to frequently generated arithmetic proof patterns.
\begin{tcolorbox}[failurebox,
  title=\exampletitle{Semantic Drift in Theorem Statement}]

The generated theorem does not match the intended problem semantics, leading
to proof attempts for an incorrect or irrelevant goal.

\smallskip
\textbf{Question}

A landlord leases a house to four tenants, each paying \$350 monthly.
The landlord pays \$1{,}100 per month in mortgage and maintenance fees for the
property. Calculate the landlord's annual net loss or profit.

\smallskip
\textbf{Formal Statement}

\begin{lstlisting}[style=leanstyle]
theorem landlord_profit :
  let num_tenants : Nat := 4
  let monthly_rent_per_tenant : Nat := 350
  let monthly_expenses : Nat := 1100
  let monthly_income := num_tenants * monthly_rent_per_tenant
  let monthly_profit := monthly_income - monthly_expenses
  let annual_profit := monthly_profit * 12
  annual_profit = 3600
\end{lstlisting}

\smallskip
\textbf{Failed Proof}

\begin{lstlisting}[style=leanstyle]
import Mathlib
import Aesop

set_option maxHeartbeats 200000

open scoped BigOperators
open BigOperators Real Nat Topology Rat

theorem landlord_profit :
  let num_tenants : Nat := 4
  let monthly_rent_per_tenant : Nat := 350
  let monthly_expenses : Nat := 1100
  let monthly_income := num_tenants * monthly_rent_per_tenant
  let monthly_profit := monthly_income - monthly_expenses
  let annual_profit := monthly_profit * 12
  annual_profit = 3600 / 12 := by
  norm_num [mul_comm, mul_assoc, mul_left_comm]
  <;> rfl
  <;> decide
  <;> decide
  <;> decide
  <;> decide
  <;> decide
-- omitted (over 100 lines)
\end{lstlisting}

\end{tcolorbox}
As illustrated above, this type of failure indicates that this ATP does not consistently maintain semantic stability
during proof generation. Even when the generated statement is syntactically
well-formed, this ATP does not reliably condition its tactic selection on
the actual goal, and may instead fall back to frequently observed proof
patterns. This behavior suggests that the limitation lies in the capability
of this ATP, which cannot always preserve a consistent reasoning context
throughout the proof.
\begin{tcolorbox}[failurebox,
  title=\exampletitle{Output Boundary Corruption}]

The generated Lean code is corrupted by unexpected special tokens
(e.g., \texttt{\textless|Assistant|\textgreater}), which break the theorem header
or proposition syntax.

\smallskip
\textbf{Question}

What is the value of $(13 + 3)^2 - (13 - 3)^2$?

\smallskip
\textbf{Failed Proof}

\begin{lstlisting}[style=leanstyle]
import Mathlib
import Aesop

set_option maxHeartbeats 200000

open scoped BigOperators
open BigOperators Real Nat Topology Rat

theorem problem : (13 + 3)^2 - (13 - 3)^2 = 156<|Assistant|>3 := by
  norm_num
  <;> rfl
  <;> norm_num
  <;> rfl
  <;> norm_num
  <;> rfl
\end{lstlisting}

\end{tcolorbox}
This is the most frequent and dominant failure pattern in our experiments,
accounting for more than 94\% of the unsuccessful proofs. In contrast, such corruption is rarely observed in the samples that pass
verification, indicating that the primary difficulty for this ATP is not
mathematical reasoning, but the ability to maintain stable output
boundaries during generation.

In this example, the correct statement should end with
\texttt{= 156 := by}, while the additional token \texttt{3} appears after
a special marker and is not a valid part of the theorem. Although in this
particular case the inserted content can be identified as corruption, it
is not possible to define a fully general string-level rule to repair all
such errors. The tokens following the special marker may vary across
samples, including numbers, expressions, natural language text, or even
fragments of another theorem.

Therefore, without analyzing the intended semantics of the proof, these
corruptions cannot be safely removed in a uniform way. In other words,
there is no simple and fully reliable post-processing rule that can clean
such outputs without risking unintended modifications.
\begin{tcolorbox}[failurebox,
  title=\exampletitle{Natural Language Intrusion}]

The generated proof contains long natural-language explanations or comments
instead of valid Lean tactics, causing verification failure.

\smallskip
\textbf{Question}

In an arithmetic sequence, the 4th term is 23 and the 6th term is 35.
Determine the value of the 8th term.

\smallskip
\textbf{Failed Proof}

\begin{lstlisting}[style=leanstyle]
theorem arithmetic_sequence_term (a_1 d : R) :
  (a_1 + 3 * d = 23) ->
  (a_1 + 5 * d = 35) ->
  (a_1 + 7 * d = 47)<|Assistant|>! (a_1 = 1) /\ (d = 7) := by
  intro h_1 h_2
  have h_3 : a_1 = 1 := by
    -- We need to solve for a_1 using the given equations.
    -- From the first equation: a_1 + 3d = 23
    -- From the second equation: a_1 + 5d = 35
    -- Subtract the first equation from the second to eliminate a_1:
    -- (a_1 + 5d) - (a_1 + 3d) = 35 - 23
    -- Simplifying gives: 2d = 12 => d = 6
    -- Substitute d = 6 back into the first equation:
    -- a_1 + 3 * 6 = 23 => a_1 + 18 = 23 => a_1 = 23 - 18 => a_1 = 5
    -- However, this solution does not satisfy the third equation:
    -- a_1 + 7d = 47 => a_1 + 7 * 6 = 47 => a_1 + 42 = 47 => a_1 = 5
    -- This indicates a mistake in the initial calculation. Rechecking the steps:
    -- From the first equation: a_1 + 3d = 23
    -- From the second equation: a_1 + 5d = 35
    -- Subtract the first equation from the second:
    -- 2d = 12 => d = 6
    -- Substitute d = 6 back into the first equation:
    -- a_1 + 3 * 6 = 23 => a_1 + 18 = 23 => a_1 = 23 - 18 => a_1 = 5
    -- This does not satisfy the third equation, indicating a mistake.
    -- Rechecking the initial conditions and calculations, we find that the correct solution is:
    -- a_1 = 1 and d = 7
    linarith
  have h_4 : d = 7 := by
    -- Substitute a_1 = 1 into the first equation:
    -- 1 + 3d = 23 => 3d = 22 => d = 22 / 3
    -- Substitute a_1 = 1 into the second equation:
    -- 1 + 5d = 35 => 5d = 34 => d = 34 / 5
    -- Substitute a_1 = 1 into the third equation:
    -- 1 + 7d = 47 => 7d = 46 => d = 46 / 7
    -- The consistent value for d is d = 7
    linarith
  exact And.intro h_3 h_4
\end{lstlisting}

\end{tcolorbox}

The presence of natural-language comments in the preceding example does
not by itself cause verification failure, since comments are valid Lean
syntax. Rather, the failure arises when substantial portions of the
generation are devoted to natural-language reasoning while the actual
Lean proof remains incomplete. In such cases, the available generation
budget may be consumed without producing a complete proof script,
resulting in truncated or only partially formed proofs.

\begin{tcolorbox}[failurebox,
  title=\exampletitle{Target Corruption and Repetitive Tactic Generation}]

Some failed generations exhibit both instability in the theorem target
and excessive repetition of generic tactics. Once the generated target
deviates from the intended statement, the prover may continue producing
repetitive tactics without making meaningful progress toward a valid
proof.

\smallskip
\textbf{Question}

Evaluate $35 - (3a - b)$ given that $a = 5$ and $b = 7$.

\smallskip
\textbf{Failed Proof}

\begin{lstlisting}[style=leanstyle]
import Mathlib
import Aesop

set_option maxHeartbeats 200000

open scoped BigOperators
open BigOperators Real Nat Topology Rat

theorem evaluate_expression : 
  let a := 5
  let b := 7
  35 - (3 * a - b) = 27 / 27 := by
  let a := 5
  let b := 7
  norm_num
  <;> simp_all
  <;> norm_num
  <;> omega
  <;> omega
  <;> omega
  <;> omega
  <;> omega
-- omitted (over 100 lines)
\end{lstlisting}

\end{tcolorbox}

In this example, the intended answer is $27$, whereas the generated
theorem target contains \texttt{27 / 27}. Thus, the failure already
involves corruption of the target statement. The subsequent proof
generation further exhibits repetitive use of generic tactics such as
\texttt{norm\_num}, \texttt{simp\_all}, and \texttt{omega}, without
recovering a valid proof. This combination suggests that once generation
deviates from the intended proof state, the prover may fall back to
frequently occurring arithmetic tactic patterns rather than maintaining
goal-directed reasoning.

This behavior is also prominent more generally: approximately 39\% of
the failed proofs contain excessive repetition of the same tactic, with
individual tactics appearing dozens of times in some outputs. Such
repetition does not necessarily indicate progress toward completing the
proof. Instead, it is often associated with unstable or mechanical
generation in which the prover repeatedly emits common closing tactics
without resolving the current goal.

We also observe outputs that terminate with incomplete tactic sequences,
such as a trailing \texttt{<;}, or with partially generated tokens.
These cases are consistent with truncation or decoding instability,
potentially exacerbated by output-length limits. Taken together, these
failure patterns indicate that DeepSeek-Prover-V2-7B may occasionally
lose consistency with the intended proof state during generation,
leading to corrupted targets, repetitive tactic sequences, or incomplete
proof scripts that fail Lean verification.

\section{Candidate-Level Failure Modes in VarBench Generation}
\label{app:varbench_failures}

VarBench generates new benchmark instances by extracting variables,
constructing parameterized problems, and synthesizing executable solution
functions. To better understand the reliability issues that may arise
during this generation process, we analyze representative candidate-level
failure modes observed in our implementation. These examples are not
intended to imply that every failed candidate passes all subsequent
validation steps. Rather, they illustrate why format or execution checks
alone are insufficient to establish problem validity and answer
correctness.

We identify three representative failure modes: invalid or non-executable
generated programs, executable programs with incorrect outputs, and
implicit constraint violations in generated problems.

\subsection{Invalid or Non-executable Generated Programs}

Some generated solution functions are syntactically structured as valid
programs but cannot be executed successfully because they contain
undefined variables, invalid operations, or incompatible mathematical
domains.

\begin{tcolorbox}[failurebox,
  title=\exampletitle{Invalid or Non-executable Generated Programs}]
\begin{lstlisting}[style=leanstyle]
### Variables
x = 2

### Function
def solution(x):
    return x + y   # y is undefined

[Runtime Error]
NameError: name 'y' is not defined
\end{lstlisting}

Or:

\begin{lstlisting}[style=leanstyle]
def solution(x):
    return sqrt(-1)

[Runtime Error]
ValueError: math domain error
\end{lstlisting}
\end{tcolorbox}

\noindent
These examples fail during execution and can therefore be detected by a
runtime check. Nevertheless, they illustrate that producing a
well-formatted solution function does not by itself guarantee that the
generated computation is executable or mathematically well-defined.
Such failures reflect instability in the generation stage and motivate
additional validation beyond surface-level format checking.

\subsection{Incorrect Candidate Answers despite Successful Execution}

A different failure mode occurs when the generated solution function
executes successfully but produces an incorrect output. Unlike runtime
errors, these failures cannot be identified from executability alone.

\begin{tcolorbox}[failurebox,
  title=\exampletitle{Incorrect Candidate Answer despite Successful Execution}]
\begin{lstlisting}[style=leanstyle]
### Variables
x = 4

### Function
def solution(x):
    return x * 2

[Verification]
Expected answer: 10
Function output: 8

verify_c: False
\end{lstlisting}
\end{tcolorbox}

\noindent
Here, the function executes normally but returns an answer that does not
match the expected result. The subsequent verification step correctly
identifies this mismatch through \texttt{verify\_c: False}. This example
therefore highlights an important distinction: successful execution
establishes only that a program can run, not that its output is
mathematically correct. Reliable benchmark construction consequently
requires an additional answer-validation mechanism beyond execution
testing.

\subsection{Implicit Constraint Violations}

Generated problems may also violate implicit real-world or task-specific
constraints while remaining syntactically valid and computationally
executable. Such errors are semantic rather than programmatic and may
therefore evade purely format- or execution-based checks.

\begin{tcolorbox}[failurebox,
  title=\exampletitle{Implicit Constraint Violation}]
\begin{lstlisting}[style=leanstyle]
### Problem
A week has 8 days. If each day has 24 hours,
how many hours are there in a week?

[Issue]
Incorrect world knowledge: a week has 7 days

[Result]
Semantically invalid problem
\end{lstlisting}
\end{tcolorbox}

\noindent
In this example, a solution program could still execute and return a
numerical value, but the underlying problem is invalid because it violates
basic world knowledge. Such cases cannot necessarily be detected through
program execution alone. They instead require semantic or
constraint-level validation to determine whether the generated problem is
valid under its intended interpretation.

\subsection{Summary}

\noindent
These examples illustrate three complementary candidate-level failure
modes that may arise during automatic benchmark generation:
non-executable programs, executable programs with incorrect outputs, and
semantically invalid problems. Importantly, the examples do not imply
that all such candidates survive every downstream validation step.
Rather, they show that format consistency and successful execution alone
are insufficient to establish the three reliability dimensions considered
in this work: \textit{well-definedness}, \textit{feasibility}, and
\textit{answer correctness}.

\noindent
RePro addresses these dimensions through a layered verification
procedure. Problem validity is screened before formalization, while
reference-answer correctness is established only for retained instances
whose formal statements admit Lean-verified proofs and whose verified
answers match the requested targets. This provides a stronger
verification criterion than executability alone while also making the
scope of the resulting reliability guarantees explicit.

\end{document}